\documentclass[sigconf]{acmart}
\usepackage{xspace}
\usepackage{soul}
\usepackage{url}
\usepackage{graphicx}
\usepackage{amsmath}
\usepackage{amsfonts}
\usepackage{multirow}
\usepackage{amsthm}
\usepackage{booktabs}
\usepackage{epstopdf}
\usepackage{xcolor}
\usepackage[labelformat=simple]{subcaption}
\usepackage{caption}
\usepackage{bbm}
\usepackage{dsfont}
\usepackage[ruled,linesnumbered]{algorithm2e}
\usepackage{setspace}
\usepackage{braket}
\usepackage{tablefootnote}
\usepackage{subcaption}
\usepackage[normalem]{ulem}
\useunder{\uline}{\ul}{}

\newcommand{\ourmethod}{GSNOP\xspace}
\graphicspath{{images/}}

\AtBeginDocument{%
  }

\setcopyright{acmcopyright}
\copyrightyear{2023}
\acmYear{2023}
\setcopyright{acmcopyright}\acmConference[WSDM '23]{Proceedings of the Sixteenth ACM International Conference on Web Search and Data Mining}{February 27-March 3, 2023}{Singapore, Singapore}
\acmBooktitle{Proceedings of the Sixteenth ACM International Conference on Web Search and Data Mining (WSDM '23), February 27-March 3, 2023, Singapore, Singapore}
\acmPrice{15.00}
\acmDOI{10.1145/3539597.3570465}
\acmISBN{978-1-4503-9407-9/23/02}

\begin{document}

\title{Graph Sequential Neural ODE Process for Link Prediction on Dynamic and Sparse Graphs}

\author{Linhao Luo}
\email{linhao.luo@monash.edu}
\affiliation{%
  \institution{Monash University}
    \country{Australia}
}

\author{Gholamreza Haffari}
\email{Gholamreza.Haffari@monash.edu}
\affiliation{%
  \institution{Monash University}
    \country{Australia}
}

\author{Shirui Pan}
\authornote{Corresponding author.}
\email{s.pan@griffith.edu.au}
\affiliation{%
  \institution{Griffith University}
  \country{Australia}
 }








\begin{abstract}
    %
    %
    
    Link prediction on dynamic graphs is an important task in graph mining. Existing approaches based on dynamic graph neural networks (DGNNs) typically require a significant amount of historical data (interactions over time), which is not always available in practice. The missing links over time, which is a common phenomenon in graph data, further aggravates the issue and thus creates extremely sparse and dynamic graphs. To address this problem, we propose a novel method based on the neural process, called Graph Sequential Neural ODE Process (\ourmethod). Specifically, \ourmethod combines the advantage of the neural process and neural ordinary differential equation (ODE) that models the link prediction on dynamic graphs as a dynamic-changing stochastic process. By defining a distribution over functions, \ourmethod introduces the uncertainty into the predictions, making it generalize to more situations instead of overfitting to the sparse data. 
    \ourmethod is also agnostic to model structures that can be integrated with any DGNN to consider the chronological and geometrical information for link prediction. Extensive experiments on three dynamic graph datasets show that \ourmethod can significantly improve the performance of existing DGNNs and outperform other neural process variants. 
\end{abstract}

\ccsdesc[500]{Networks~Network algorithms}
\ccsdesc[500]{Computing methodologies~Machine learning}

\keywords{Link Prediction, Dynamic Graphs, Neural Process, Graph Neural Networks, Neural Ordinary Differential Equation}

\maketitle

\section{Introduction}\label{sec:intro}
\begin{figure}[t]
    \centering
    \includegraphics[width=1\columnwidth,trim=0 0 0cm 0, clip]{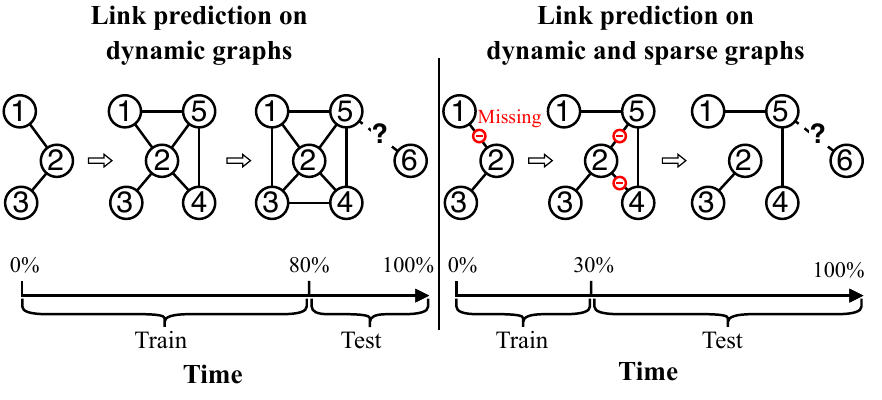}
    \caption{Comparison between link prediction on dynamic graphs and link prediction on dynamic and sparse graphs.}
    \label{fig:sparse_dyg}
\end{figure}

Graphs are ubiquitous data structures in modeling the interactions between objects. As a fundamental task in graph data mining, link prediction has been found in many real-world applications, such as social media analysis \cite{huo2018link}, drug discovery \cite{abbas2021application}, and recommender systems \cite{6413904}. However, data in these applications is often dynamic. For example, people interact with their friends on social media every day, and customers contiguously click or buy products in a recommender system. Therefore, it is essential to consider the nature of dynamics for accurate link prediction. As shown in Figure \ref{fig:sparse_dyg}, link prediction on \textit{dynamic graphs}, where interactions arrive over time, aims to predict links in the future by using historical interactions \cite{sankar2020dysat,kumar2019predicting,jin2022neural}.

Recently, graph neural networks (GNNs) have shown great capacity in mining graph data \cite{welling2016semi,luo2021detecting,liu2021anomaly,zheng2022multi,xiong2022pseudo,lin2022distributed}. Under the umbrella of dynamic graph neural networks (DGNNs), several approaches have recently been proposed for predicting links on dynamic graphs \cite{wang2021inductive,tgn_icml_grl2020,wang2021apan}. As shown in the left panel of Figure \ref{fig:sparse_dyg}, these methods typically require a large amount of historical links for training, which is unfortunately not always available \cite{yang2022few,zheng2022rethink}. 
For instance, a new-start online shopping platform could only have few days' user-item interactions. DGNNs trained on such limited data would be easily overfitting and cannot generalize to new links. In contrast, the platform expects models to accurately predict future interactions as early as possible, which leaves little time to collect the data. Furthermore, because the links in dynamic graphs arrive over time, the limitation of historical data in dynamic graphs would be exacerbated by the missing of interactions, resulting in extremely sparse graphs. Thus, how to enable effective link prediction on \textit{dynamic} and \textit{sparse} graphs (e.g., with limited historical links, as shown in the right panel of Figure \ref{fig:sparse_dyg}) remains a significant challenge in this area.

Neural process (NP) \cite{garnelo2018neural}, a new family of methods, opens up a new door to dealing with limited data points in machine learning 
\cite{singh2019sequential,norcliffe2020neural,nguyen2022transformer}.  Based on the stochastic process, NP first draws a distribution of functions from the data by using an aggregator, and then it samples a function from the learned distribution for predictions. By defining the distribution, NP can not only readily adapt to newly observed data, but also estimate the uncertainty over the predictions, which makes models generalize to more situations, rather than overfitting to the sparse data. NP has already been applied in many applications, such as 3D rendering \cite{eslami2018neural}, cold-start recommendation \cite{lin2021task}, and link prediction on static graphs \cite{liang2022neural}.


\begin{figure}
    \centering
    \newlength{\MyHeight}
    \settoheight{\MyHeight}{\includegraphics[width=0.49\columnwidth]{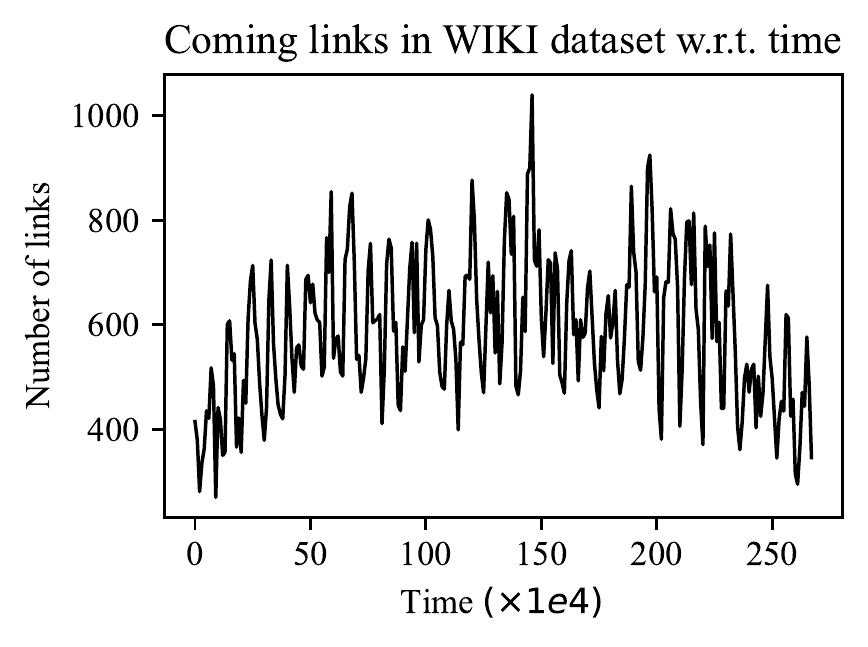}}
        \begin{subfigure}[t]{0.44\columnwidth}
            \centering
            \begin{minipage}[b][\MyHeight][c]{1\columnwidth}
                \includegraphics[width=1\textwidth,trim=0 0.7cm 1.3cm 0, clip]{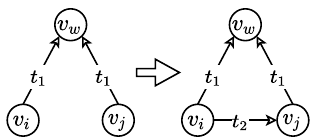}
            \end{minipage}
            \caption{Sequential and structural dependence.}
            \label{subfig:dependence}
        \end{subfigure}
        \hfill
        \begin{subfigure}[t]{0.55\columnwidth}
            \centering
            \includegraphics[width=1\textwidth]{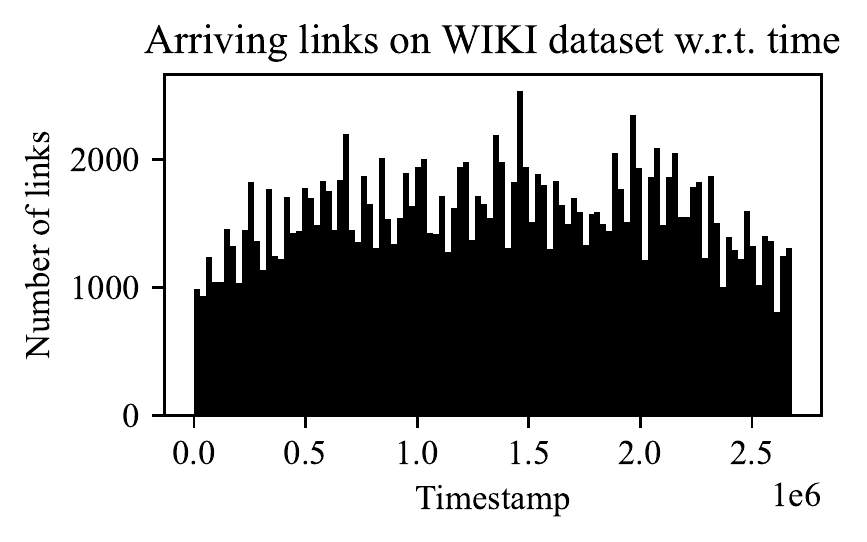}
            \caption{Irregular arriving links in dynamic graphs.}
            \label{subfig:irregular}
        \end{subfigure}
        \caption{(a) Sequential and structural dependence  and (b) irregular arriving links in dynamic graphs.}
        \label{fig:intro}
\end{figure}

However, NP and its variants \cite{kim2018attentive,singh2019sequential,cangea2022message} cannot be directly applied to the link prediction on the dynamic graphs. Vanilla NP \cite{garnelo2018neural} draws the distribution using a simple average-pooling aggregator, which ignores the sequential and structural dependence between nodes for accurate link prediction \cite{wang2021inductive,kovanen2011temporal,paranjape2017motifs}. For example, as shown in the Figure \ref{subfig:dependence}, node $v_i$ and $v_j$ respectively have an interaction with node $v_w$ at time $t_1$. Therefore, node $v_i$ and $v_j$ are quite likely to interact in the future $t_2$. In this case, simply using average-pooling aggregator would erase the dependence and lead to a sub-optimal result. SNP \cite{singh2019sequential} adopts a RNN aggregator to consider the sequential information, but it still fails to consider the derivative of the underlying distribution and uses a static distribution through time.
For instance, as shown in Figure \ref{subfig:irregular}, the links in dynamic graphs arrive irregularly \cite{cao2021inductive,han2021learning,wen2022trend}. Additionally, an important event could lead to a large number of links occurring in a short time, which is shown by the spikes in Figure \ref{subfig:irregular}. Therefore, using a static distribution drawn from the historical data is inappropriate.

In this paper, we introduce a novel class of NP, called Graph Sequential Neural ODE Process (\ourmethod) for link prediction on dynamic graphs. Specifically, \ourmethod first draws a distribution of link prediction functions from historical data using a \textit{dynamic graph neural network encoder}, which can be incorporated with any existing DGNN model to capture the chronological and geometrical information of dynamic graphs. Second, we propose a \textit{sequential ODE aggregator}, in which we consider the function of dynamic graph link prediction as a sequence of dynamic-changing stochastic processes, and we adopt the RNN model to specify the dependence between data. To handle the irregular data, we apply the neural ordinary differential equation (ODE) \cite{chen2018neural} which is a continuous-time model defining the derivative of the hidden state with a neural network. Neural ODE is used to model the derivative of the neural process and infer the distribution in the future. In this way, we can obtain the distribution of link predict functions at any contiguous timestamp. Finally, we sample the function from the distribution for link prediction, which introduces the uncertainty over prediction results, offering a resilient way to deal with the sparse and limited data.

The main contributions of this paper are summarized as follows:
\begin{itemize}
    \item We propose a novel class of NP, called Graph Sequential Neural ODE Process (\ourmethod). To the best of our knowledge, this is the first work to apply NP for link prediction on dynamic and sparse graphs.
    \item We propose a dynamic graph neural network encoder and a sequential ODE aggregator, which inherits the merits of neural process and neural ODE to model the dynamic-changing stochastic process.
    \item We conduct extensive experiments on three public dynamic graph datasets. Experiment results show that \ourmethod can significantly improve the performance of existing DGNNs on dynamic and sparse graphs.
\end{itemize}
\section{Related work}
In this section, we first review some representative DGNN models and then introduce the NP and its variants.

\subsection{Dynamic Graph Neural Networks}\label{sec:dgnn}
Dynamic graph neural networks (DGNNs) can be roughly grouped into two categories based on the type of dynamic graphs: discrete-time dynamic graphs (DTDG) and continuous-time dynamic graphs (CTDG). DTDG-based methods model dynamic graphs as a sequence of graph snapshots. DySAT \cite{sankar2020dysat} first generates node embeddings in each snapshot using an attention-based GNN, then summarize the embeddings with a Temporal Self-Attention. EvloveGCN \cite{pareja2020evolvegcn} first trains a GNN model on each snapshot and adopts the RNN to evolve the parameters of GNN to capture the temporal dynamics. Continuous-time dynamic graph (CTDG) is a more general case where links arrive with contiguous timestamps. JODIE \cite{kumar2019predicting} employs two recurrent neural networks to update the node embedding by message passing. TGAT \cite{xu2020inductive} proposes a temporal graph attention layer to aggregate temporal-topological neighbors. 
Furthermore, TGN \cite{tgn_icml_grl2020} introduces a memory module to incorporate the historical interactions. APAN \cite{wang2021apan} decouples the message passing and memory updating process, which greatly improves the computational speed in CTDG. However, these DGNNs require sufficient training data, which largely limits their applications. MetaDyGNN \cite{yang2022few} adopts the meta-learning framework for few-shot link prediction on dynamic graphs. But it requires new observed links to finetune the model, which shares different settings with our work.

\subsection{Neural Process}
Conditional neural process (CNP) \cite{garnelo2018conditional} is the first member of the NP family. CNP encodes the data into a deterministic hidden variable that parametrizes the function, which does not introduce any uncertainty. To address the limitation of CNP, neural process (NP) \cite{garnelo2018neural} is a stochastic process that learns a latent variable to model an underlying distribution over functions, from which we can sample a function for downstream tasks. ANP \cite{kim2018attentive} marries the merits of CNP and NP by incorporating the deterministic and stochastic paths in an attentive way. However, ANP could be memory-consuming when processing a long sequence, since it needs to perform self-attention on all the data points.
SNP \cite{singh2019sequential} further adopts RNN to consider the sequential information by modeling the changing of distribution. But it fixes the distribution in testing due to the lack of new observed data. NDP \cite{norcliffe2020neural} applies the neural ODE \cite{chen2018neural} to the decoder of neural process to consider the dynamic between data.
NPs have also been applied to address many problems, such as modeling stochastic physics fields \cite{holderrieth2021equivariant}, node classification \cite{cangea2022message}, and link prediction on static graphs \cite{liang2022neural}. However, none of them applies NP for dynamic graph link prediction. Besides, existing NPs cannot be directly used in dynamic graphs as they ignore the sequential and structural dependence as well as the irregular data.
\section{Problem Definition and Preliminary}
In this section, we formally define our problem and introduce some key concepts used in this paper.

\subsection{Problem Definition}
Dynamic graphs can be grouped into two categories: discrete-time dynamic graphs (DTDGs) and continuous-time dynamic graphs (CTDGs). A DTDG can be defined as a sequence of graph snapshots separated by time interval: $\mathcal{G}=\{\mathcal{G}_1,\mathcal{G}_2,\ldots,\mathcal{G}_T\}$, where $ \mathcal{G}_t=(\mathcal{V}_t,\mathcal{E}_t)$ is a graph taken at a discrete timestamp $t$. A CTDG is a more general case of a dynamic graph, which is constituted as a sequence of links stream $e(t)=(v_i,v_j,t)$ coming in over time, where $e(t)$ denotes node $v_i$ and $v_j$ has an interaction at a continuous timestamp $t$. Thus, CTDG can be formulated as $\mathcal G=\{e(t_1),e(t_2),\ldots\}$.

\noindent\textbf{Definition 1: Link Prediction on Dynamic Graphs.}
Given a CTDG $\mathcal{G}$, our task is to predict links $e(t)$ at future timestamp $t> T$, using limited historical interactions $\mathcal{G}_{\leq T}=\{e(t_1),e(t_2),\ldots,e(t_n)\}$, where $t_i\in[1,T]$.

\subsection{Preliminary}

\subsubsection{Neural Ordinary Differential Equation}
Neural ordinary differential equation (ODE) \cite{chen2018neural} is a continuous-time model which defines the derivative of the hidden state with a neural network. This can be formulated as
\begin{equation}
    \setlength\abovedisplayskip{2pt}
    \setlength\belowdisplayskip{2pt}
    \frac{dz(t)}{dt} = f_{ode}(z(t),t),
\end{equation}
where $z(t)$ denotes the hidden state of a dynamic system at time $t$, and $f_{ode}$ denotes a neural network that describe the derivative of the hidden state w.r.t. time. Thus, the neural ODE can output the $z(t)$ at any timestamp by solving the following equation
\begin{equation}
    \setlength\abovedisplayskip{2pt}
    \setlength\belowdisplayskip{2pt}
    z(t) = z(t_0) + \int_{t_0}^{t} f_{ode}(z(t_i),t)dt_i.\label{eq:init_node}
\end{equation}
Equation \ref{eq:init_node} can be solved by using numerical ODE solver, such as Euler and fourth-order Runge-Kutta. Thus, it can be simplified as 
\begin{equation}
    \setlength\abovedisplayskip{2pt}
    \setlength\belowdisplayskip{2pt}
    z(t) = ODESolve(f_{ode},z(t_0),t_0,t).
\end{equation}

\subsubsection{Neural Process}
Neural process (NP) \cite{garnelo2018neural,garnelo2018conditional} is a stochastic process that learns an underlying distribution of the prediction function $f: X\to Y$ with limited training data. Specifically, NP defines a latent variable model using a set of \textit{context data} $C=\{(x_i,y_i)\}_{i=1}^n$ to model a prior distribution $P(z|C)$. Then, the prediction likelihood on the \textit{target data} $D=\{(x_i,y_i)\}_{i=n}^{n+m}$ is modeled as 
\begin{equation}
    \setlength\abovedisplayskip{2pt}
    \setlength\belowdisplayskip{2pt}
    P(y_{n:n+m}|x_{n:n+m},C)=\int_{z} P(y_{n:n+m}|x_{n:n+m},z)P(z|C)dz.
\end{equation}

\begin{figure}[t]
    \centering
    \includegraphics[width=0.8\columnwidth,trim=0 0 0cm 0, clip]{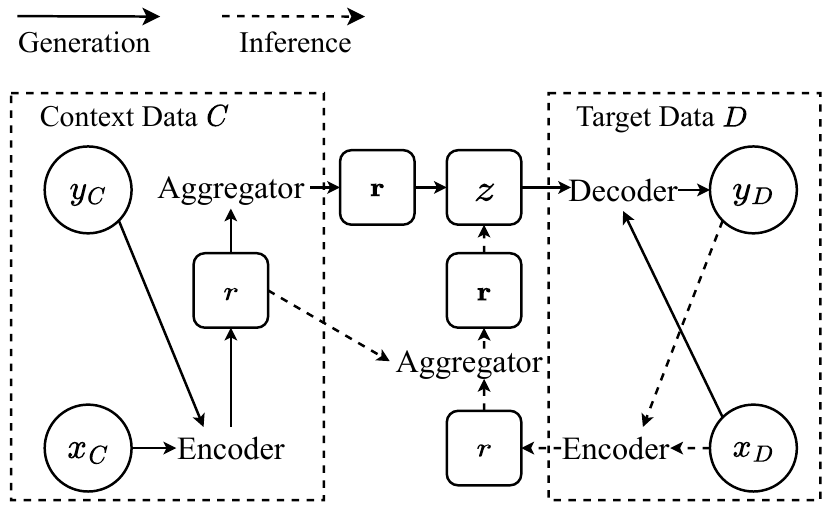}
    \caption{The framework of neural process. Circles denote the input and output data; round squares denote the latent variables. }
    \label{fig:np_framework}
\end{figure}

We illustrate the framework of NP in Figure \ref{fig:np_framework}. The aim of NP is to infer the target distribution from context data. 
NP first draws a distribution of $f$ from the context data $C$, written as $P_\theta(z|C)$. By using an \textit{encoder}, NP learns a low-dimension vector $r$ for each pair $(x_C, y_C)$ in the context data. Then, it adopts an \textit{aggregator} to synthesize all the $r$ into a global representation $\mathbf{r}$, which parametrizes the distribution, formulated as $z\sim\mathcal{N}\big(\mu(\mathbf{r}),\sigma(\mathbf{r})\big)$. Last, it samples a $z$ from the distribution which is fed together with the $x_D\in D$ into a \textit{decoder} to predict the label for target data, which is written as $P(y_{n:n+m}|x_{n:n+m},z)$. Because the true posterior is intractable, the NP is trained via variational approximation by maximizing the evidence lower bound (ELBO) as follows
\begin{equation}
    \setlength\abovedisplayskip{2pt}
    \setlength\belowdisplayskip{2pt}
    \begin{split}
        &log P(y_{n:n+m}|x_{n:n+m},C)\geq\\
        &\mathbb{E}_{Q_\psi(z|C,D)}[logP_\phi(y_{n:n+m}|x_{n:n+m},z)]-KL\big(Q_\psi(z|C,D)||P_\theta(z|C)\big),
    \end{split}
\end{equation}
where $Q_\psi(z|C,D)$ approximates the true posterior distribution also parameterized by a neural network.

\begin{figure*}[t]
    \centering
    \includegraphics[width=0.65\textwidth,trim=0 0 0cm 0, clip]{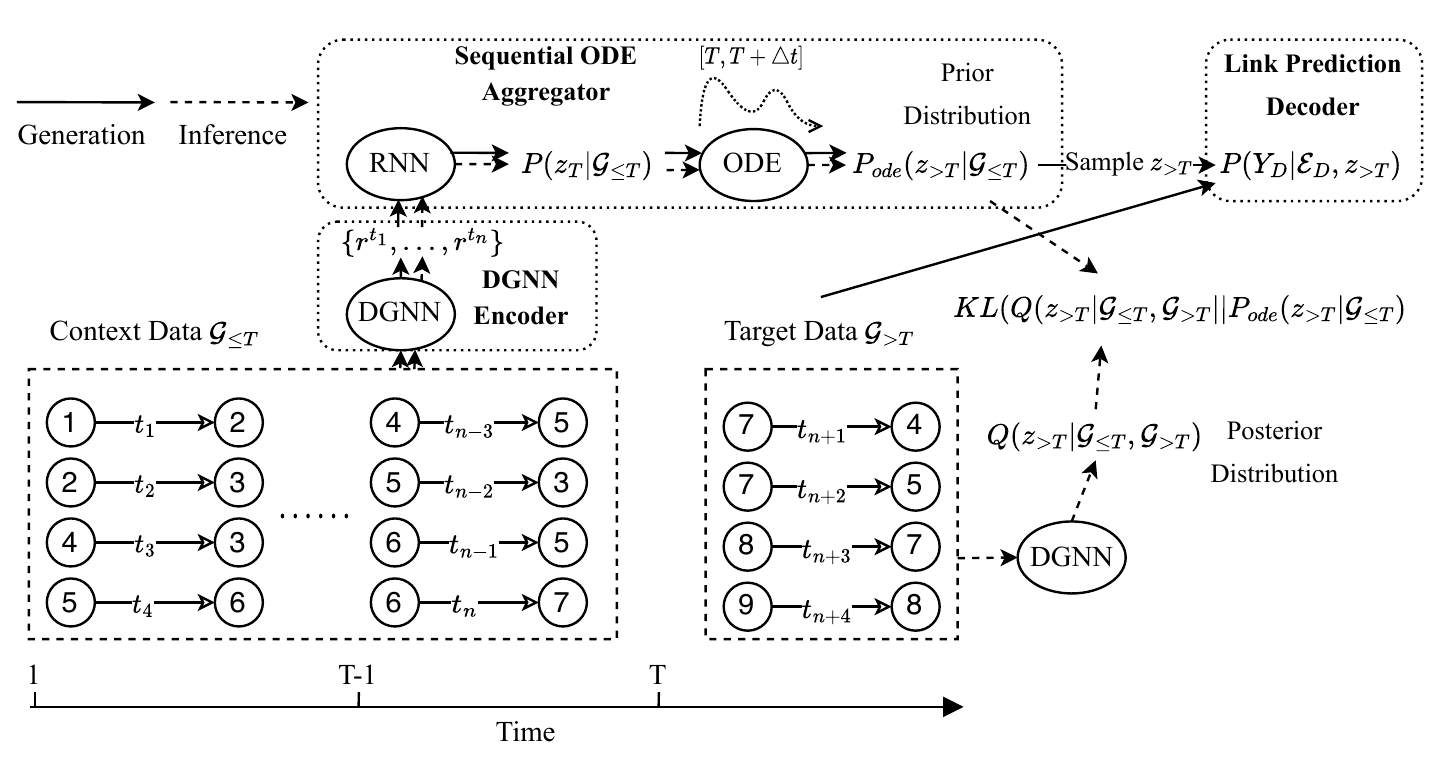}
    \caption{The framework of Graph Sequential Neural ODE Process (\ourmethod) for link prediction on dynamic and sparse graphs.}
    \label{fig:framework}
\end{figure*}

\section{Graph Sequential Neural ODE Process}\label{sec:gsonp}
In this section, we introduce the proposed Graph Sequential Neural ODE Process (\ourmethod), which combines the merits of neural process and neural ODE to model the dynamic-changing stochastic process. \ourmethod can be integrated with any dynamic graph neural network (DGNN) to simultaneously consider the chronological and geometrical information for dynamic graph link prediction. The framework of \ourmethod is illustrated in Figure \ref{fig:framework}.

\subsection{Setup and Framework}
Given a CTDG $\mathcal{G}$, we adopt \ourmethod to infer the distribution of an underlying link prediction function $f: (e(t))\to y, y\in\{0,1\}$ from context data.
By specifying a timestamp $T$, we split $\mathcal{G}$ into the context data $\mathcal{G}_{\leq T}$ and target data $\mathcal{G}_{>T}$. Each $e_C(t)\in \mathcal{G}_{\leq T}$ occurred prior to $T$, and $e_D(t)\in \mathcal{G}_{> T}$ arrives when $t> T$. At testing time, we treat all the historical interactions as context data, and predict links in the future.
\ourmethod follows the framework of NP, thus, the \textit{encoder}, \textit{aggregator}, and \textit{decoder} in \ourmethod are defined as follows.

As shown in Figure \ref{fig:framework}, \ourmethod first draws a historical distribution of $f$ from the context data $\mathcal{G}_{\leq T}$ by using a \textit{dynamic graph neural network encoder}, written as $P(z_{T}|\mathcal{G}_{\leq T})$. \ourmethod adopts the dynamic graph neural network to learn a low-dimension vector $r^t$ for each link in the context data. Then, it adopts an \textit{sequential ODE aggregator} to synthesize all the $r^t$ into a global representation $\mathbf{r}_{T}$, which parametrizes the distribution $P(z_{T}|\mathcal{G}_{\leq T})$, formulated as $z_{T}\sim\mathcal{N}\big(\mu(\mathbf{r}_{T}),\sigma(\mathbf{r}_{T})\big)$. Following, \ourmethod applies the neural ODE on the $P(z_{T}|\mathcal{G}_{\leq T})$ to infer the distribution in the future $>T$, written as $P_{ode}(z_{>T}|\mathcal{G}_{\leq T})$
Last, \ourmethod sampled a $z_{>T}$ from the future distribution which is fed together with the $e_D(t)\in \mathcal{G}_{>T}$ into a \textit{link prediction decoder} to predict the links in target data, which is formulated as $f(e_D(t),z_{>T})\to y_T$. Each component is introduced in detail as follows.

\subsection{Dynamic Graph Neural Network Encoder}\label{sec:DGNN}

The encoder is applied to each pair $(e_C(t), y_C)$ in the context data, where $y_C$ denotes the existence of the links. It tries to capture the connection between $e_C(t)$ and $y_C$ to infer the distribution of link prediction function $f$. Link prediction on dynamic graph has been studied for years. Previous research has shown that the dynamic graph neural networks (DGNNs) have great potential to capture the chronological and geometrical information for dynamic graph link prediction \cite{xu2020inductive,tgn_icml_grl2020}. Therefore, we adopt the DGNNs as the encoder of \ourmethod.

Given a pair $(e_C(t), y_C)$, we first adopt a DGNN to generate the representation $h^t$ of nodes $v_i,v_j\in e_C(t)$, which is formulated as 
\begin{equation}
    \setlength\abovedisplayskip{2pt}
    \setlength\belowdisplayskip{2pt}
    \label{eq:DGNN}
    h^t = \text{DGNN}\Big(h^{t-1},agg\big(N_{< t}(v)\big)\Big),
\end{equation}
where $h^{t-1}$ denotes the representation of node $v$ in the last timestamp, $N_{< t}(v)$ denotes a set of nodes that interacted with $v$ in previous timestamps, and $agg(\cdot)$ denotes the aggregation operation defined by DGNN. Equation \ref{eq:DGNN} can be written as $\text{DGNN}(v,N_{< t}(v))$ for simplification. We can use arbitrary DGNN to consider the dynamic and graph structure information for a given node. 

Then, we use a MLP to encode each pair $(e_C(t), y_C)$ into a representation $r^t$, which is formulated as 
\begin{equation}
    \setlength\abovedisplayskip{2pt}
    \setlength\belowdisplayskip{2pt}
    r^t = \text{MLP}\Big(h_i^t||h_j^t||y\Big) + t_{emb}, y=\left\{
        \begin{aligned}
            1, e_C(t) \in \mathcal{G}_{\leq T}\\
            0, e_C(t) \notin \mathcal{G}_{\leq T}
        \end{aligned}
    \right.,
    \label{eq:encoder}
\end{equation}
where $||$ denotes the concatenation operation, $y$ denotes the existence of the link between $v_i$ and $v_j$ in the context data, and $t_{emb}$ denotes the time embedding \cite{xu2020inductive} sharing the same dimension with $r^t $ that can be learned during the training. 

\subsection{Sequential ODE Aggregator}\label{sec:aggregator}
The aggregator summarizes encoder outputs into a global representation $\mathbf{r}_{T}$ that parametrizes latent distribution $z_{T}\sim \mathcal{N}\big(\mu(\mathbf{r}_{T}), \sigma(\mathbf{r}_{T})\big)$. Originally, NP \cite{chen2018neural} adopts a simple average-pooling aggregator, which is formulated as
\begin{equation}
    \setlength\abovedisplayskip{2pt}
    \setlength\belowdisplayskip{2pt}
    \mathbf{r}_{T} = \frac{1}{|\mathcal{G}_{\leq T}|}\sum_{e_C(t)\in \mathcal{G}_{\leq T}}r^t.
    \label{eq:aggregator}
\end{equation}
However, it fails to consider the \textit{sequential} and \textit{structural} dependence between nodes in dynamic graphs. 

Vanilla NP assumes that data in the context data is i.i.d., thus it adopts a simple average-pooling aggregator to define the joint distribution $P(z_{T}|\mathcal{G}_{\leq T})$. However, in dynamic graphs, we need to consider the sequential and structural dependence between nodes for accurate link prediction \cite{wang2021inductive,kovanen2011temporal,paranjape2017motifs}.
In this case, simply using average-pooling aggregator would erase the dependence between $e_C(t)\in \mathcal{G}_{\leq T}$ and lead to a sub-optimal result.

To consider the sequential and structural dependence, inspired by Sequential Neural Process (SNP) \cite{singh2019sequential}, we consider the dynamic graph link prediction function as a sequence of dynamic-changing stochastic processes $\{\mathcal{P}_1,\ldots,\mathcal{P}_T\}$. At each timestamp $t$, we can observe a set of context data $\{(e(t), y)\in \mathcal{G}_{t}\}$, where $\mathcal{G}_{t}$ might differ over time. We can infer a distribution $\mathcal{P}_t$ by using all the historical data $\mathcal{G}_{\leq t}$, which can be formulated as
\begin{align}
    \mathcal{P}_t&=P(z_t|\mathcal{G}_{\leq t}) =P(z_t|\mathcal{P}_{t-1},\mathcal{G}_t).\label{eq:snp}
\end{align}
Because, $\mathcal{P}_{t-1}$ denotes the distribution drawn from the $\mathcal{G}_{<t}$. By modeling the changing of process $\mathcal{P}_{t-1}\to \mathcal{P}_t$, we can consider the dependence among context data. Thus, we adopt a RNN model to specify the state transition model in Eq. (\ref{eq:snp}), which is formulated as 
\begin{equation}
    \mathbf{r}_{t} = \text{RNN}(\mathbf{r}_{t-1}, \mathcal{G}_t), t > 1,\label{eq:rnn}
\end{equation}
where $\mathbf{r}_1$ is computed by averaging $r^1$ generated from $\mathcal{G}_1$.
By using the RNN aggregator, we can obtain the $\mathbf{r}_{T}$ from the representation of context data $\{r^{t_1},\ldots,r^{T}\}$ and define the distribution $P(z_{T}|\mathcal{G}_{\leq T})$. But, it still fails to consider the derivative of the underlying distribution.

In vanilla NP, it infers the distribution of link prediction function $P(z_{T}|\mathcal{G}_{\leq T})$ from historical context data, and fixes it for future link prediction. However, in dynamic graphs, the links arrive in different frequencies \cite{cao2021inductive,han2021learning,wen2022trend}. 
Therefore, it is inappropriate to use a static distribution over time, although we can incorporate the new-coming links into the context data and update the distribution by using Eq. \ref{eq:rnn}. In most cases, it is infeasible to obtain the links in the future. As a result, we need to consider the derivative of distribution over time to deal with irregular data.


Recently, neural ordinary differential equation (ODE) \cite{chen2018neural} has shown great power in dealing irregular time series data \cite{rubanova2019latent}. Thus, in \ourmethod, we apply the neural ODE to model the derivative of neural process and infer the distribution in the future $P_{ode}(z_{>T}|\mathcal{G}_{\leq T})$.

Given a distribution $P(z_{T}|\mathcal{G}_{\leq T})$ drawn from context data $\mathcal{G}_{\leq T}$, it is parameterized by the global latent variable $\mathbf{r}_{T}$. We apply the neural ODE on the $\mathbf{r}_{T}$ to obtain the latent variable at any future time $>T=T+\triangle t$, which is formulated as
\begin{equation}
    \setlength\abovedisplayskip{2pt}
    \setlength\belowdisplayskip{2pt}
    \mathbf{r}_{>T} = \mathbf{r}_{T} + \int_{T}^{T+\triangle t} f_{ode}(\mathbf{r}_t, t) dt,\label{eq:ode}
\end{equation}
where $f_{ode}$ denotes the neural ODE function, which is formulated as
\begin{equation}
    \setlength\abovedisplayskip{2pt}
    \setlength\belowdisplayskip{2pt}
    f_\theta(\mathbf{r}_t, t) = Tanh(MLP(\mathbf{r}_t + t_{emb})).\label{eq:node}
\end{equation}
We use the $Tanh$ as the activation function to keep from gradient exploration.

After obtaining the $r_{>T}$, the distribution parameterized by $\mathbf{r}_{>T}$ can be considered as a Gaussian distribution $\mathcal{N}\big(\mu(\mathbf{r}_{>T}),\sigma(\mathbf{r}_{>T})\big)$, where the mean $\mu(\mathbf{r}_{>T})$ and variance $\sigma(\mathbf{r}_{>T})$ are formulated as 
\begin{gather}
    \chi = \text{MLP}(\mathbf{r}_{>T}),\\
    \mu(\mathbf{r}_{>T}) = ReLU(\text{MLP}(\chi)),\\
    \sigma(\mathbf{r}_{>T}) = 0.1 + 0.9*Sigmoid(\text{MLP}(\chi)).
\end{gather}

\subsection{Link Prediction Decoder}\label{sec:dec}
The decoder is considered as the underlying link prediction function $f(e(t),z)\to y$. Given two nodes $v_i$ and $v_j$ at time $t$ to be predicted, $f$ can be formulated as 
\begin{gather}
    \setlength\abovedisplayskip{2pt}
    \setlength\belowdisplayskip{2pt}
    h_i^t = \text{DGNN}(v_i,N_{<t}(v_i)),h_j^t = \text{DGNN}(v_j,N_{<t}(v_i))\\
    \text{Sample } z_{>T}\sim\mathcal{N}\big(\mu(\mathbf{r}_{>T}),\sigma(\mathbf{r}_{>T})\big)\\
    \widetilde{h_i^{t}} = ReLU(\text{MLP}(h_j^t||z_{>T})), \widetilde{h_j^{t}} = ReLU(\text{MLP}(h_j^t||z_{>T})),\\
    \hat{y} = Sigmoid\big(\text{MLP}(\widetilde{h_i^{t}}||\widetilde{h_j^{t}})\big),
\end{gather}
where $\hat{y}$ denotes the prediction result. 

The decoder is conditioned on the variable $z_{>T}$ sampled from the distribution. This injects the uncertainty into the prediction results, making \ourmethod generalize to more situations, even with limited data. In addition, as more data observed, the distribution could be refined and provide more accurate predictions, which is demonstrated in section \ref{sec:sparsity}.  

\subsection{Optimization}

Given the context data $\mathcal{G}_{\leq T}$ and target data $\mathcal{G}_{> T}$, the object of \ourmethod is to infer the distribution $P(z_{>T}|\mathcal{G}_{\leq T})$ from the context data that minimize the prediction loss on the target data. Therefore, the generation process of \ourmethod can be written as 
\begin{equation}
    P(Y_D,z_{>T}|\mathcal{E}_D,\mathcal{G}_{\leq T}) = P(z_{>T}|\mathcal{G}_{\leq T}) \prod_{e_D(t)\in \mathcal{E}_D} P\big(y_D|f(e_D(t),z_{>T})\big),
\end{equation}
where $f$ is the link prediction decoder, and $\mathcal{E}_D=\{e_D(t)\in \mathcal{G}_{>T}\}$ denotes the links to be predicted in the target data.

Since the decoder $f$ is non-linear, it can be trained using an amortised variational inference procedure. We can optimize the evidence lower bound (ELBO) to minimize the prediction loss on target data given the context data $logP(Y_D|\mathcal{E}_D,\mathcal{G}_{\leq T})$, which can be derivate as
\begin{align}
    \setlength\abovedisplayskip{2pt}
    \setlength\belowdisplayskip{2pt}
    &logP(Y_D|\mathcal{E}_D,\mathcal{G}_{\leq T}) = \int_{z}Q(z_{>T})log\frac{P(Y_D, z_{>T}|\mathcal{E}_D,\mathcal{G}_{\leq T})}{P(z_{>T}|\mathcal{G}_{\leq T})},\\
    &\geq \int_{z}Q(z_{>T})log\frac{P(Y_D, z_{>T}|\mathcal{E}_D,\mathcal{G}_{\leq T})}{Q(z_{>T})}, \\
    &= \mathbb{E}_{Q(z_{>T})}log\frac{P(Y_D, z_{>T}|\mathcal{E}_D,\mathcal{G}_{\leq T})}{Q(z_{>T})},\\
    &= \mathbb{E}_{Q(z_{>T})}[logP(Y_D|\mathcal{E}_D,z_{>T}) + log\frac{P(z_{>T}|\mathcal{G}_{\leq T})}{Q(z_{>T})}],\\
    &= \mathbb{E}_{Q(z_{>T})}[logP(Y_D|\mathcal{E}_D,z_{>T})] 
    - KL\big(Q(z_{>T})||P(z_{>T}|\mathcal{G}_{\leq T})\big),
\end{align}
where $Q(z_{>T})$ denotes the true posterior distribution of $z_{>T}$.

Thus, we can obtain the final ELBO loss as
\begin{equation}
    \begin{split}
        \mathcal{L}_{ELBO} = &\mathbb{E}_{Q_\psi(z_{>T}|\mathcal{G}_{\leq T},\mathcal{G}_{> T})}[log P_\phi(Y_D|\mathcal{E}_D,z_{>T})]\\
        &- KL\big(Q_\psi(z_{>T}|\mathcal{G}_{\leq T},\mathcal{G}_{> T})||P_{ode}(z_{>T}|\mathcal{G}_{\leq T})\big).
    \end{split}
    \label{eq:elbo}
\end{equation}
where $log P_\phi(Y_D|\mathcal{E}_D,z_{>T}) = \sum_{e_D(t)\in \mathcal{E}_D}P_\phi(y_D|e_D(t),z_{>T})$ denotes the prediction loss on the target data and $P_\phi$ is parameterized by the decoder. The $Q(z_{>T})$ is approximated by the variational posterior $Q_\psi(z_{>T}|\mathcal{G}_{\leq T},\mathcal{G}_{> T})$ calculated by the encoder and RNN aggregator, formulated as
\begin{gather}
        r^{t} = \text{DGNN Encoder}\big((e_D(t),y_D)\big), (e_D(t),y_D) \in \mathcal{G}_{> T},\\ 
        \mathbf{r'}_{>T} = \text{RNN}(\mathbf{r}_{T},r^{t}),\\
        Q_\psi(z_{>T}|\mathcal{G}_{\leq T},\mathcal{G}_{> T}) = \mathcal{N}\big(\mu(\mathbf{r'}_{>T}),\sigma(\mathbf{r'}_{>T})\big).\label{eq:pos}
\end{gather}

The prior $P(z_{>T}|\mathcal{G}_{\leq T})$ is approximated by the neural ODE in Eq. \ref{eq:ode}, written as $P_{ode}(z_{>T}|\mathcal{G}_{\leq T})$. By minimizing the KL divergence between variational posterior and the prior $P_{ode}(z_{>T}|\mathcal{G}_{\leq T})$, it encourages the neural ODE to learn the derivative of distribution and infer the distribution in the future. Extended derivation of Eq. \ref{eq:elbo} can be found in the supplementary file of this link\footnotemark[2].

Noticeably, unlike neural ODE process \cite{norcliffe2020neural} which applies neural ODE on the decoder, \ourmethod applies neural ODE to infer the distribution of function at any contiguous timestamp. This is more suitable for the dynamic-changing stochastic process modeled for dynamic graph link prediction.


\noindent\textbf{Testing.}
At test time, given a set of historical interactions $\mathcal{G}_{\leq T}$, we treat them as the context data and draw a global representation $\mathbf{r}_{T}$ using the DGNN encoder as well as RNN aggregator. Then, given a link $e(T+\triangle t)$ to be predicted, we apply the neural ODE to infer the distribution at future, formulated as
\begin{gather}
    \setlength\abovedisplayskip{2pt}
    \setlength\belowdisplayskip{2pt}
    r_{>T}=ODESolve(f_{ode},r_T,T, T+\triangle t),\\
    \text{Sample } z_{>T}\sim\mathcal{N}\big(\mu(\mathbf{r}_{>T}),\sigma(\mathbf{r}_{>T})\big).
\end{gather}
Last, we use the decoder to predict the link $f(e(T+\triangle t),z_{>T})\to y$.

Given a CTDG containing $n$ links in context data and $m$ links to be predicted in the target data, the time complexity of \ourmethod is $O(n+m)$ since model need to encode each link in the context data and predict each target link. This shows that \ourmethod scales well with the size of the input. At testing time, we also batch the samples by using the dummy variable trick \cite{chen2020neural} to speed up the integration in neural ODE.


\section{Experiment}

\subsection{Datasets}
We conduct experiments on three widely used CTDG datasets: WIKI \cite{tgn_icml_grl2020}, REDDIT \cite{tgn_icml_grl2020}, and MOOC \cite{kumar2019predicting}.
Following the settings of the previous work \cite{xu2020inductive}, we focus on the task of predicting links in the future using limited historical interactions. Intuitively, with the links coming over time, the CDTG would become denser.
To simulate different levels of graph sparsity, we split the CDTG chronologically by timestamps in the entire duration with two ratios, 30\%-20\%-50\% and 10\%-10\%-80\% for training, validation, and testing. The datasets with different splits are labeled by their training ratio (e.g., WIKI\_0.3 and WIKI\_0.1). The sparsity of graphs can be quantified by the density score, calculated as $\frac{2|\mathcal{E}|}{|\mathcal{V}|(|\mathcal{V}|-1)}$, where $|\mathcal{E}|$ and $|\mathcal{V}|$ denote the number of links and nodes in the training set. The statistics of the datasets is shown in Table \ref{tab:dataset}.

\begin{table}[htbp]
    \caption[Dataset statistics. The max($t$) column shows the maximum timestamps of links. The $d_e$ denotes the dimension of link features. Density (0.3) and Density (0.1) respectively denote the sparsity of CDTG under different splits.]{Dataset statistics. The max($t$) column shows the maximum timestamps of links. The $d_e$ denotes the dimension of link features\footnotemark. Density (0.3) and Density (0.1) respectively denote the density scores of CDTG under different splits.}
    \label{tab:dataset}
    \resizebox{\columnwidth}{!}{%
        \begin{tabular}{@{}ccccccc@{}}
            \toprule
            Dataset & $|\mathcal{V}_{all}|$  & $|\mathcal{E}_{all}|$   & max($t$) & $d_e$ & Density (0.3) & Density (0.1) \\ \midrule
            WIKI    & 9,228  & 157,474 & 2.70E+06 & 172   & 1.03E-03       & 2.90E-04       \\
            REDDIT  & 10,985 & 672,447 & 2.70E+06 & 172   & 3.18E-03       & 1.01E-03       \\
            MOOC    & 7,047  & 411,749 & 2.60E+06 & 128   & 4.03E-03       & 1.52E-03       \\ \bottomrule
        \end{tabular}%
    }
\end{table}
\footnotetext{Due to the lack of link features, we randomly generated link features for MOOC.}

\subsection{Baselines}

Since \ourmethod is agnostic to model structure, we choose several state-of-the-art DGNN as baselines (i.e., JODIE \cite{kumar2019predicting}, DySAT \cite{sankar2020dysat}, TGAT \cite{xu2020inductive}, TGN \cite{tgn_icml_grl2020}, and APAN \cite{wang2021apan}). Detailed introductions of these DGNNs can be found in section \ref{sec:dgnn}.
To demonstrate the effectiveness of \ourmethod, we adopt these DGNNs respectively as the encoder in \ourmethod to illustrate the performance improved by \ourmethod. In addition, we also select other neural process variants (i.e., Neural Process (NP) \cite{garnelo2018neural}, Conditional Neural Process \cite{garnelo2018conditional} and Sequential Neural Process (SNP) \cite{singh2019sequential}) to demonstrate the superiority of \ourmethod.

\subsection{Settings}
In experiments, we adopt the average precision (AP) and mean reciprocal rank (MRR) as evaluation metrics. To fairly evaluate the performance in sparse settings \cite{yokoi2017link}, we randomly select 50 negative links for each positive link in the evaluation stage.

For each DGNN baseline, we set the layer of the GNN to 2 and the number of neighbors per layer to 10. The memory size for APAN is set to 10, while 1 for JODIE and TGN. For snapshot-based DySAT, we set the number of snapshots to 3 with the duration of each snapshot to be 10,000 seconds. For all methods, the dropout rate is swept from \{0.1,0.2,0.3,0.4,0.5\}, and the dimension of output node embeddings is set to 100. We use the function in section \ref{sec:dec} to predict links.

For \ourmethod, the dimension of $r^t$, $t_{emb}$, and $z$ are all set to 256. The Monte Carlo sampling size for approximating $\mathbb{E}_{Q_\psi(z_{>T}|G_{\leq T},G_{> T})}$ in $\mathcal{L}_{ELBO}$ is set to 10. We use GRU \cite{chung2014empirical} as the RNN aggregator and choose 5 order Runge-Kutta of Dormand-Prince-Shampine \cite{calvo1990fifth} as the ODE solver with relative tolerance set to $1e-5$ and absolute tolerance set to $1e-7$. We use Adam as the optimizer, and the learning rate is set to 0.00001. The implementation of \ourmethod is available at this link \footnote{\url{https://github.com/RManLuo/GSNOP}}.

\begin{table*}[htbp]
    \centering
    \caption{Experiments results on WIKI, REDDIT, and MOOC datasets.}
    \label{tab:comparsion}
    \resizebox{0.75\textwidth}{!}{%
    \begin{tabular}{@{}c|cccccc|cccccc@{}}
        \toprule
        \multirow{2}{*}{Method} & \multicolumn{2}{c}{WIKI\_0.3} & \multicolumn{2}{c}{REDDIT\_0.3} & \multicolumn{2}{c|}{MOOC\_0.3} & \multicolumn{2}{c}{WIKI\_0.1} & \multicolumn{2}{c}{REDDIT\_0.1} & \multicolumn{2}{c}{MOOC\_0.1}                                                                                                             \\ \cmidrule(l){2-13}
                                & AP                            & MRR                             & AP                            & MRR                           & AP                              & MRR                           & AP              & MRR             & AP              & MRR             & AP              & MRR             \\ \midrule
        JODIE                   & 0.3329                        & 0.5595                          & 0.6320                        & 0.8101                        & 0.6690                          & 0.8802                        & 0.1820          & 0.4476          & 0.3324          & 0.6401          & 0.7100          & 0.9063          \\
        JODIE+\ourmethod        & \textbf{0.4258}               & \textbf{0.6086}                 & \textbf{0.8171}               & \textbf{0.8922}               & \textbf{0.6716}                 & \textbf{0.8853}               & \textbf{0.2979} & \textbf{0.5273} & \textbf{0.4705} & \textbf{0.7260} & \textbf{0.7196} & \textbf{0.9066} \\\midrule
        DySAT                   & 0.3881                        & 0.6680                          & 0.5705                        & 0.7887                        & 0.6441                          & 0.8807                        & 0.3872          & 0.6601          & 0.5353          & 0.7853          & 0.6705          & 0.8968          \\
        DySAT+\ourmethod        & \textbf{0.5816}               & \textbf{0.7716}                 & \textbf{0.7406}               & \textbf{0.8443}               & \textbf{0.6528}                 & \textbf{0.8811}               & \textbf{0.3910} & \textbf{0.6628} & \textbf{0.5738} & \textbf{0.7887} & \textbf{0.6718} & \textbf{0.8971} \\\midrule
        TGAT                    & 0.3253                        & 0.5229                          & 0.8343                        & 0.8797                        & 0.5183                          & 0.7828                        & 0.2766          & 0.4638          & 0.4833          & 0.6523          & \textbf{0.1919}          & 0.4228          \\
        TGAT+\ourmethod         & \textbf{0.3701}               & \textbf{0.5348}                 & \textbf{0.8395}               & \textbf{0.8820}               & \textbf{0.5448}                 & \textbf{0.7905}               & \textbf{0.3366} & \textbf{0.4961} & \textbf{0.5240} & \textbf{0.6807} & 0.1874 & \textbf{0.4581} \\\midrule
        TGN                     & 0.5541                        & 0.7788                          & 0.7621                        & 0.8587                        & 0.7200                          & 0.9007                        & 0.5676          & 0.7631          & 0.6749          & 0.8029          & 0.7677          & 0.9087          \\
        TGN+\ourmethod          & \textbf{0.6633}               & \textbf{0.7816}                 & \textbf{0.8307}               & \textbf{0.8935}               & \textbf{0.7445}                 & \textbf{0.9117}               & \textbf{0.6568} & \textbf{0.7674} & \textbf{0.7148} & \textbf{0.8263} & \textbf{0.7714} & \textbf{0.9174} \\\midrule
        APAN                    & 0.2431                        & 0.5496                          & 0.6166                        & 0.7799                        & 0.6601                          & 0.8774                        & 0.0504          & 0.1857          & 0.5601          & 0.7415          & 0.7016          & 0.8967          \\
        APAN+\ourmethod         & \textbf{0.4570}               & \textbf{0.6918}                 & \textbf{0.7119}               & \textbf{0.8560}               & \textbf{0.6614}                 & \textbf{0.8790}               & \textbf{0.1996} & \textbf{0.5607} & \textbf{0.6126} & \textbf{0.7606} & \textbf{0.7052} & \textbf{0.8982} \\
        \bottomrule
    \end{tabular}%
    }
\end{table*}

\begin{table*}[htbp]
    \centering
    \caption{Experiments results on WIKI and REDDIT w.r.t. different sample ratios.}\label{tab:sample_ratio}
    \resizebox{0.85\linewidth}{!}{%
        \begin{tabular}{c|cccccccccc|cccccccccc}
            \toprule
            Dataset          & \multicolumn{10}{c|}{WIKI\_0.3} & \multicolumn{10}{c}{REDDIT\_0.3}                                                                                                                                                                                                                                                                                                                                                                                                               \\
            \midrule
            Sample Ratio     & \multicolumn{2}{c}{100\%}       & \multicolumn{2}{c}{80\%}         & \multicolumn{2}{c}{50\%} & \multicolumn{2}{c}{20\%} & \multicolumn{2}{c|}{10\%} & \multicolumn{2}{c}{100\%} & \multicolumn{2}{c}{80\%} & \multicolumn{2}{c}{50\%} & \multicolumn{2}{c}{20\%} & \multicolumn{2}{c}{10\%}                                                                                                                                                                                     \\
            \midrule
            Method           & AP                              & MRR                              & AP                       & MRR                      & AP                        & MRR                       & AP                       & MRR                      & AP                       & MRR                      & AP              & MRR             & AP              & MRR             & AP              & MRR             & AP              & MRR             & AP              & MRR             \\
            \midrule
            JODIE            & 0.3329                          & 0.5595                           & 0.2804                   & 0.5046                   & 0.1464                    & 0.3950                    & 0.0300                   & 0.1598                   & 0.0238                   & 0.1289                   & 0.1820          & 0.4476          & 0.1966          & 0.4574          & 0.1215          & 0.3959          & 0.0222          & 0.1207          & 0.0246          & 0.1366          \\
            JODIE+\ourmethod & \textbf{0.4258}                 & \textbf{0.6086}                  & \textbf{0.3177}          & \textbf{0.5171}          & \textbf{0.2666}           & \textbf{0.4738}           & \textbf{0.1845}          & \textbf{0.4192}          & \textbf{0.0663}          & \textbf{0.2784}          & \textbf{0.2979} & \textbf{0.5273} & \textbf{0.2382} & \textbf{0.4816} & \textbf{0.1824} & \textbf{0.4157} & \textbf{0.1017} & \textbf{0.3302} & \textbf{0.0406} & \textbf{0.2214} \\\midrule
            DySAT            & 0.3881                          & 0.6680                           & 0.3856                   & 0.6691                   & 0.3873                    & 0.6676                    & 0.3870                   & 0.6610                   & 0.3121                   & 0.6403                   & 0.3872          & 0.6601          & 0.3179          & 0.6244          & 0.3165          & 0.6385          & 0.3161          & 0.6450          & 0.3118          & 0.6413          \\
            DySAT+\ourmethod & \textbf{0.5816}                 & \textbf{0.7716}                  & \textbf{0.5539}          & \textbf{0.6691}          & \textbf{0.4349}           & \textbf{0.4349}           & \textbf{0.3900}          & \textbf{0.6615}          & \textbf{0.3159}          & \textbf{0.6419}          & \textbf{0.3910} & \textbf{0.6628} & \textbf{0.3870} & \textbf{0.6612} & \textbf{0.3174} & \textbf{0.6441} & \textbf{0.3229} & \textbf{0.6534} & \textbf{0.3050} & \textbf{0.6409} \\
            \midrule
            TGAT             & 0.3253                          & 0.5229                           & 0.3274                   & 0.5238                   & 0.2717                    & 0.4769                    & 0.2573                   & 0.4783                   & 0.2536                   & 0.4722                   & 0.2766          & 0.4638          & 0.2782          & 0.4861          & 0.1215          & 0.3830          & 0.1356          & 0.3545          & 0.1114          & 0.3567          \\
            TGAT+\ourmethod  & \textbf{0.3701}                 & \textbf{0.5348}                  & \textbf{0.3594}          & \textbf{0.5281}          & \textbf{0.3630}           & \textbf{0.5314}           & \textbf{0.3371}          & \textbf{0.4955}          & \textbf{0.2882}          & \textbf{0.4571}          & \textbf{0.3366} & \textbf{0.4961} & \textbf{0.3420} & \textbf{0.5055} & \textbf{0.3055} & \textbf{0.4586} & \textbf{0.2045} & \textbf{0.3731} & \textbf{0.0600} & \textbf{0.2408} \\
            \midrule
            TGN              & 0.5541                          & 0.7788                           & 0.5327                   & 0.7560                   & 0.5084                    & 0.7341                    & 0.4361                   & 0.6809                   & 0.4918                   & 0.6831                   & 0.5676          & 0.7631          & 0.3576          & 0.6632          & 0.3633          & 0.6576          & 0.4160          & 0.6604          & 0.0312          & 0.1681          \\
            TGN+\ourmethod   & \textbf{0.6633}                 & \textbf{0.7816}                  & \textbf{0.6395}          & \textbf{0.7762}          & \textbf{0.6179}           & \textbf{0.7685}           & \textbf{0.5535}          & \textbf{0.7094}          & \textbf{0.5566}          & \textbf{0.6968}          & \textbf{0.6568} & \textbf{0.7674} & \textbf{0.6435} & \textbf{0.7679} & \textbf{0.5506} & \textbf{0.7330} & \textbf{0.5094} & \textbf{0.6868} & \textbf{0.3974} & \textbf{0.6374} \\
            \midrule
            APAN             & 0.2431                          & 0.5496                           & 0.2198                   & 0.5374                   & 0.1768                    & 0.5179                    & 0.0928                   & 0.2524                   & 0.0431                   & 0.1969                   & 0.0504          & 0.1857          & 0.0547          & 0.1900          & 0.0631          & 0.2442          & 0.0367          & 0.1388          & 0.0331          & 0.1226          \\
            APAN+\ourmethod  & \textbf{0.4570}                 & \textbf{0.6918}                  & \textbf{0.4363}          & \textbf{0.6821}          & \textbf{0.2068}           & \textbf{0.5311}           & \textbf{0.0735}          & \textbf{0.2797}          & \textbf{0.0588}          & \textbf{0.2128}          & \textbf{0.1996} & \textbf{0.5607} & \textbf{0.0638} & \textbf{0.1958} & \textbf{0.0644} & \textbf{0.3075} & \textbf{0.0430} & \textbf{0.1964} & \textbf{0.0288} & \textbf{0.1173} \\
            \midrule\midrule
            Dataset          & \multicolumn{10}{c|}{WIKI\_0.1} & \multicolumn{10}{c}{REDDIT\_0.1}                                                                                                                                                                                                                                                                                                                                                                                                               \\
            \midrule
            Sample Ratio     & \multicolumn{2}{c}{100\%}       & \multicolumn{2}{c}{80\%}         & \multicolumn{2}{c}{50\%} & \multicolumn{2}{c}{20\%} & \multicolumn{2}{c|}{10\%} & \multicolumn{2}{c}{100\%} & \multicolumn{2}{c}{80\%} & \multicolumn{2}{c}{50\%} & \multicolumn{2}{c}{20\%} & \multicolumn{2}{c}{10\%}                                                                                                                                                                                     \\
            \midrule
            Method           & AP                              & MRR                              & AP                       & MRR                      & AP                        & MRR                       & AP                       & MRR                      & AP                       & MRR                      & AP              & MRR             & AP              & MRR             & AP              & MRR             & AP              & MRR             & AP              & MRR             \\
            \midrule
            JODIE            & 0.6320                          & 0.8101                           & 0.6118                   & 0.8085                   & 0.4374                    & 0.7306                    & 0.3391                   & 0.6597                   & 0.2984                   & 0.5949                   & 0.3324          & 0.6401          & 0.3183          & 0.6307          & 0.3134          & 0.6179          & 0.3156          & 0.6226          & 0.0380          & 0.1854          \\
            JODIE+\ourmethod & \textbf{0.8171}                 & \textbf{0.8922}                  & \textbf{0.7543}          & \textbf{0.8617}          & \textbf{0.4860}           & \textbf{0.7546}           & \textbf{0.3907}          & \textbf{0.6806}          & \textbf{0.3130}          & \textbf{0.6140}          & \textbf{0.4705} & \textbf{0.7260} & \textbf{0.4428} & \textbf{0.6653} & \textbf{0.3637} & \textbf{0.6585} & \textbf{0.3760} & \textbf{0.6681} & \textbf{0.0635} & \textbf{0.2436} \\\midrule
            DySAT            & 0.5705                          & 0.7887                           & 0.5714                   & 0.7888                   & 0.5732                    & 0.7925                    & 0.5786                   & 0.7919                   & 0.5751                   & 0.7872                   & 0.5353          & 0.7853          & 0.5458          & 0.7795          & 0.5371          & 0.7845          & 0.5568          & 0.7818          & 0.5598          & 0.7798          \\
            DySAT+\ourmethod & \textbf{0.7406}                 & \textbf{0.8443}                  & \textbf{0.7281}          & \textbf{0.8390}          & \textbf{0.6106}           & \textbf{0.7977}           & \textbf{0.6166}          & \textbf{0.7982}          & \textbf{0.6110}          & \textbf{0.7964}          & \textbf{0.5738} & \textbf{0.7887} & \textbf{0.5755} & \textbf{0.7945} & \textbf{0.5721} & \textbf{0.7955} & \textbf{0.5786} & \textbf{0.7954} & \textbf{0.5947} & \textbf{0.7942} \\
            \midrule
            TGAT             & 0.8343                          & 0.8797                           & 0.8193                   & 0.8702                   & 0.8074                    & 0.8604                    & 0.7775                   & 0.8490                   & 0.4192                   & 0.6160                   & 0.4833          & 0.6523          & 0.4885          & 0.6512          & 0.3783          & 0.5651          & 0.2129          & 0.4615          & 0.2056          & 0.4518          \\
            TGAT+\ourmethod  & \textbf{0.8395}                 & \textbf{0.8820}                  & \textbf{0.8426}          & \textbf{0.8839}          & \textbf{0.8168}           & \textbf{0.8757}           & \textbf{0.7917}          & \textbf{0.8544}          & \textbf{0.4513}          & \textbf{0.6295}          & \textbf{0.5240} & \textbf{0.6807} & \textbf{0.5247} & \textbf{0.6862} & \textbf{0.4181} & \textbf{0.6041} & \textbf{0.3709} & \textbf{0.5667} & \textbf{0.2426} & \textbf{0.4684} \\
            \midrule
            TGN              & 0.7621                          & 0.8587                           & 0.7118                   & 0.8471                   & 0.7247                    & 0.8424                    & 0.3549                   & 0.6572                   & 0.3005                   & 0.5934                   & 0.6749          & 0.8029          & 0.6631          & 0.7917          & 0.5958          & 0.7502          & 0.2403          & 0.5312          & 0.2219          & 0.5152          \\
            TGN+\ourmethod   & \textbf{0.8307}                 & \textbf{0.8935}                  & \textbf{0.8235}          & \textbf{0.8967}          & \textbf{0.7567}           & \textbf{0.8444}           & \textbf{0.5923}          & \textbf{0.7941}          & \textbf{0.5024}          & \textbf{0.7373}          & \textbf{0.7148} & \textbf{0.8263} & \textbf{0.6995} & \textbf{0.8095} & \textbf{0.5134} & \textbf{0.7494} & \textbf{0.2701} & \textbf{0.5458} & \textbf{0.1929} & \textbf{0.4410} \\
            \midrule
            APAN             & 0.6166                          & 0.7799                           & 0.4879                   & 0.7527                   & 0.4914                    & 0.7306                    & 0.4637                   & 0.7294                   & 0.4943                   & 0.7148                   & 0.5601          & 0.7415          & 0.5733          & 0.7407          & 0.2991          & 0.6973          & 0.3566          & 0.6761          & 0.3586          & 0.6677          \\
            APAN+\ourmethod  & \textbf{0.7119}                 & \textbf{0.8560}                  & \textbf{0.7229}          & \textbf{0.8498}          & \textbf{0.5370}           & \textbf{0.7500}           & \textbf{0.4865}          & \textbf{0.7331}          & \textbf{0.4925}          & \textbf{0.7133}          & \textbf{0.6126} & \textbf{0.7606} & \textbf{0.6131} & \textbf{0.7418} & \textbf{0.4157} & \textbf{0.7403} & \textbf{0.4397} & \textbf{0.6831} & \textbf{0.3756} & \textbf{0.6823} \\
            \bottomrule
        \end{tabular}
    }
\end{table*}

\subsection{Performance Comparison}
In this section, we evaluate the performance of \ourmethod on different datasets. The results are shown in Table \ref{tab:comparsion}. From the results, we can see that \ourmethod enables to improve the performance of all the baselines with respect to different datasets and training ratios.

Specifically, when the historical training data is sufficient, all DGNNs achieve relatively good performance in WIKI\_0.3, REDDIT\_0.3, and MOOC\_0.3. However, when training data is limited, the performance of some baselines drops significantly (e.g., JODIE, TGAT, and APAN). The possible reason is that these DGNNs use historical data to generate node embeddings. When data is limited, the quality of the generated embeddings cannot be guaranteed. Thus, it is inaccurate to use such embeddings for link prediction directly. In addition, the data distribution of the testing set could be diverse from the training set. This would lead the model overfit to the historical data and cannot generalize to the future links. The performances between MOOC\_0.3 and MOOC\_0.1 are similar. This is because MOOC is a denser CTDG with short duration. Thus, most DGNNs perform well in this dataset.

In \ourmethod, we model the distribution of function from the historical data by learning a latent variable. This variable introduces a global view of knowledge, from which we can sample a link prediction function rather than only use the node embeddings. Besides, the distribution introduces uncertainty over prediction results. In this way, \ourmethod can generalize to more situations instead of overfitting to the training data. Last, by adopting the neural ODE, \ourmethod enables to infer the distribution in the future with a small amount of historical data. Noticeably, the improvement of \ourmethod is sometimes limited (e.g., results in MOOC\_0.1). The possible reason is that \ourmethod learns the latent variable by optimizing the evidence lower bound. This does not always result in a meaningful latent variable that parametrizes the distribution \cite{alemi2018fixing,nguyen2022transformer}. We will leave this for future exploration.

\subsection{Performance at Different Levels of Sparsity}\label{sec:sparsity}
In the real-world, we might only observe partial links for an underlying CTDG, which aggravates the sparsity. To further demonstrate the performance improvement under different sparsity, we sample subsets of training links from the original training set with various sample ratios, while leave the validation and test set unchanged. For instance, the 30\% sample ratio in WIKI\_0.3 denotes that we randomly mask 70\% of links in the training set and use the remaining 30\% links for training. The 
density score of the sampled dataset can be calculated by multiplying the sample ratio and the density of the original dataset. We sweep the sample ratio from 100\% to 10\%.
The results are shown in Table \ref{tab:sample_ratio}.

\begin{table*}[tbp]
    \centering
    \caption{Comparing with other neural process variants.}
    \label{tab:abiliation}
    \resizebox{.95\linewidth}{!}{%
        \begin{tabular}{c|cccccccccc|cccccccccc}
            \toprule
            \multirow{3}{*}{Method} & \multicolumn{10}{c|}{WIKI\_0.3} & \multicolumn{10}{c}{WIKI\_0.1}                                                                                                                                                                                                                                                                                                                                                \\
            \cmidrule{2-21}
                                    & \multicolumn{2}{c}{Origin}      & \multicolumn{2}{c}{NP}         & \multicolumn{2}{c}{CNP} & \multicolumn{2}{c}{SNP} & \multicolumn{2}{c|}{\ourmethod} & \multicolumn{2}{c}{origin} & \multicolumn{2}{c}{NP} & \multicolumn{2}{c}{CNP} & \multicolumn{2}{c}{SNP} & \multicolumn{2}{c}{\ourmethod}                                                                                                               \\
            \cmidrule{2-21}
                                    & AP                              & MRR                            & AP                      & MRR                     & AP                              & MRR                        & AP                     & MRR                     & AP                      & MRR                            & AP     & MRR    & AP     & MRR    & AP      & MRR     & AP     & MRR    & AP              & MRR             \\
            \midrule
            JODIE                   & 0.3868                          & 0.6241                         & 0.4063                  & 0.6118                  & 0.3255                          & 0.5943                     & 0.4173                 & 0.6023                  & \textbf{0.4258}         & \textbf{0.6086}                & 0.1946 & 0.4741 & 0.2223 & 0.4699 & 0.1201  & 0.3609  & 0.2804 & 0.5149 & \textbf{0.2979} & \textbf{0.5273} \\
            DySAT                   & 0.3757                          & 0.6727                         & 0.4226                  & 0.6818                  & 0.3494                          & 0.6552                     & 0.4199                 & 0.6821                  & \textbf{0.5816}         & \textbf{0.7716}                & 0.3152 & 0.6461 & 0.3151 & 0.6452 & 0.3482  & 0.6386  & 0.3874 & 0.6612 & \textbf{0.3910} & \textbf{0.6628} \\
            TGAT                    & 0.3277                          & 0.5171                         & 0.3456                  & 0.5147                  & 0.2881                          & 0.4680                     & 0.3462                 & 0.5173                  & \textbf{0.3701}         & \textbf{0.5348}                & 0.2766 & 0.4638 & 0.3051 & 0.4704 & 0.2407 & 0.4474 & 0.2996 & 0.4646 & \textbf{0.3366} & \textbf{0.4961} \\
            TGN                     & 0.5611                          & 0.7789                         & 0.6632                  & 0.7837                  & 0.6539                          & 0.7766                     & 0.6577                 & 0.7809                  & \textbf{0.6633}         & \textbf{0.7816}                & 0.5778 & 0.7750 & 0.6362 & 0.7617 & 0.5703  & 0.7347  & 0.6479 & 0.7656 & \textbf{0.6568} & \textbf{0.7674} \\
            APAN                    & 0.2191                          & 0.5085                         & 0.0784                  & 0.2701                  & 0.0987                          & 0.3195                     & 0.1090                 & 0.3503                  & \textbf{0.4570}         & \textbf{0.6918}                & 0.1598 & 0.5251 & 0.0820 & 0.3115 & 0.0836  & 0.2837  & 0.1135 & 0.3995 & \textbf{0.1996} & \textbf{0.5607} \\
            \bottomrule
        \end{tabular}
    }
\end{table*}

From the results, we can see that \ourmethod outperforms the baselines under different levels of sparsity. Specifically, with the sample ratio decreasing, the performance of each DGNN also decreases. This is because when the training data is sparse, the model cannot capture enough interactions for link prediction. However, with \ourmethod, we can define the distribution with limited interactions to estimate the underlying graph. Thus, it can better handle the sparsity situation.
In addition, \ourmethod can incorporate new observations to enhance the distribution. With the sample ratio increasing from 10\%-100\%, the improvement bought by \ourmethod increases. The reason is that by adding new observed links into the latent variable, \ourmethod enables to more accurately define the distribution of function with the help of DGNN encoder and RNN aggregator.

\subsection{Ablation Study}
In this section, we compare our \ourmethod with other neural process variants (i.e., vanilla neural process (NP) \cite{garnelo2018neural}, conditional neural process \cite{garnelo2018conditional} and sequential neural process (SNP) \cite{singh2019sequential}) to show the superiority of our method in dynamic graph link prediction. For each variant, we use our dynamic graph neural network encoder and link prediction decoder but maintain their original way of generating the distribution. 
The results in Table \ref{tab:abiliation} would help to demonstrate the effectiveness of our sequential ODE aggregator.

From the results, we can see that \ourmethod performs better than other neural process variants. NP is the original version of neural process. Since it ignores the sequential and structural dependence, NP has the minimum improvement. Compared with other variants, CNP achieves the worst performance. The reason is that CNP is a deterministic version of neural process. Instead of learning the latent variable, CNP directly models the distribution conditioned on the context data, which could be inaccurate when the context data is limited. SNP adopts the RNN as the aggregator, which considers the sequential dependence. However, it fails to capture the distribution derivative, making SNP unable to handle irregular data. \ourmethod incorporates both the RNN aggregator and neural ODE, which simultaneously exploits the sequential information and enables inferring the distribution at any timestamps. Thus, \ourmethod achieves the best results among all the variants.


\begin{figure}[tbp]
    \centering
    \newlength{\subMyHeight}
    \settoheight{\subMyHeight}{\includegraphics[width=0.4\columnwidth]{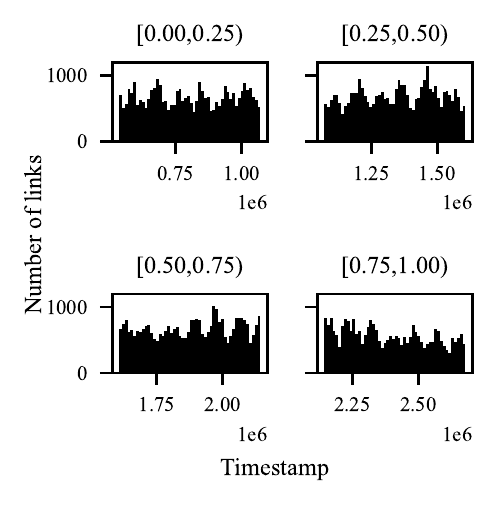}}
        \begin{subfigure}[t]{0.49\columnwidth}
            \centering
            \begin{minipage}[b][\subMyHeight][c]{1\columnwidth}
            \includegraphics[width=1\textwidth]{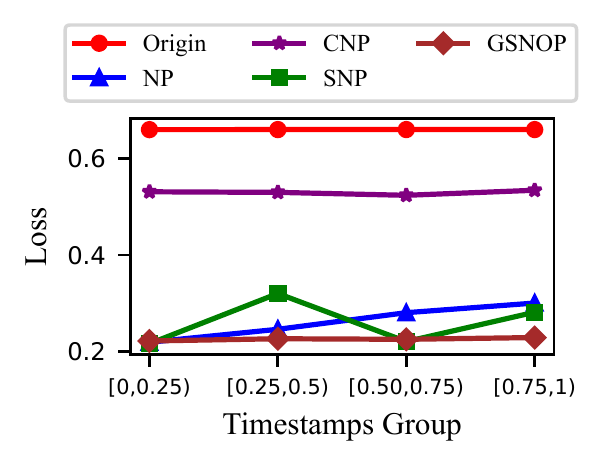}
            \end{minipage}
            \caption{Prediction loss of each group.}
            \label{subfig:loss}
        \end{subfigure}
        \hfill
        \begin{subfigure}[t]{0.4\columnwidth}
            \centering
                \includegraphics[width=1\textwidth,trim=0 0.0cm 0cm 0, clip]{images/wiki_edges_group_hist.pdf}
            \caption{Arriving links of each group.}
            \label{subfig:groups}
        \end{subfigure}
        \caption{(a) Prediction loss and (b) arriving links of each group.}
\end{figure}

To further demonstrate the effectiveness of Sequential ODE aggregator, we split links in the test set of WIKI\_0.1 chronologically into four groups based on their timestamps. For example, the group [0, 0.25) denotes the links between $0$ and $0.25 * max(t)$ in the test set. The prediction loss of each group with DySAT as the encoder is shown in Figure \ref{subfig:loss}. From the results, we can see that Origin and CNP have relatively high losses across different groups. Due to the ignoring of sequential dependence, NP cannot infer the distribution in the future. Thus, the loss of NP increases with the timestamps. Despite SNP capturing the sequential information, it still fails to address the irregular data. As shown in Figure \ref{subfig:groups}, there is a high spike in group [0.25, 0.5), and the arriving links are decreasing in group [0.75, 1.00). Thus, the losses of SNP are higher in these irregular groups. With the help of the sequential ODE aggregator, \ourmethod models the distribution derivative and maintain relatively low losses across different timestamps. 

\subsection{Parameters Analysis}\label{sec:parameters}
\begin{figure}[tbp]
    \centering
    \includegraphics[width=0.9\columnwidth,trim=0 0cm 0cm 0, clip]{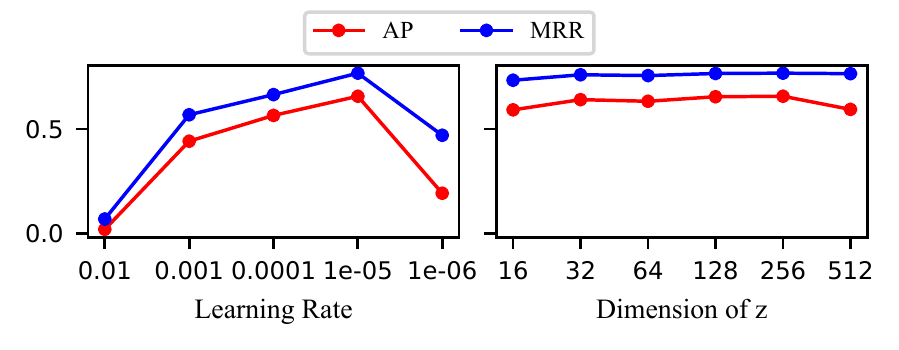}
    \caption{Parameter sensitivity w.r.t. different learning rate and the dimension of $z$.}
    \label{fig:parameter}
\end{figure}

We investigate two critical hyperparameters (i.e., learning rate and the dimension of $z$) in Figure \ref{fig:parameter}. From the results, we can see that the learning rate could severely impact the performance of \ourmethod. This is because \ourmethod adopts the dummy variable trick \cite{chen2020neural} to speed up the neural ODE. However, this will incur the gradient explosion problem when integrating in a large time period. Thus, we need to select a relatively small learning rate (i.e., $1e-5$) to avoid gradient explosion. However, a too small learning rate also causes underfitting and leads the performance drops. With the dimension of $z$ increasing, the performance of \ourmethod improves and reaches the best at 256. The reason is that a small dimension of $z$ cannot well model the stochastic of distribution, but an over large dimension would also introduce noises when optimizing the evidence lower bound, which deteriorates the performance.

\section{Conclusion}
In this paper, we propose a novel NP named \ourmethod for link prediction on dynamic graphs. We first specify a dynamic graph neural network encoder to capture the chronological and geometrical information of dynamic graphs. Then, we propose a sequential ODE aggregator, which not only considers the sequential dependence among data but also learns the derivate of the neural process. In this way, it can better draw the distribution from limited historical links. We conduct extensive experiments on three public datasets, and the results show that \ourmethod can significantly improve the link prediction performance of DGNNs in dynamic and sparse graphs. Experiment results comparing with other NP variants further demonstrate the effectiveness of our method. 

\section{Acknowledge}
This work is supported by an ARC Future Fellowship (No. FT21010 0097).
\clearpage
\balance
\bibliographystyle{ACM-Reference-Format}
\bibliography{sections/ref.bib}


\begin{thebibliography}{42}


\ifx \showCODEN    \undefined \def \showCODEN     #1{\unskip}     \fi
\ifx \showDOI      \undefined \def \showDOI       #1{#1}\fi
\ifx \showISBNx    \undefined \def \showISBNx     #1{\unskip}     \fi
\ifx \showISBNxiii \undefined \def \showISBNxiii  #1{\unskip}     \fi
\ifx \showISSN     \undefined \def \showISSN      #1{\unskip}     \fi
\ifx \showLCCN     \undefined \def \showLCCN      #1{\unskip}     \fi
\ifx \shownote     \undefined \def \shownote      #1{#1}          \fi
\ifx \showarticletitle \undefined \def \showarticletitle #1{#1}   \fi
\ifx \showURL      \undefined \def \showURL       {\relax}        \fi
\providecommand\bibfield[2]{#2}
\providecommand\bibinfo[2]{#2}
\providecommand\natexlab[1]{#1}
\providecommand\showeprint[2][]{arXiv:#2}

\bibitem[Abbas et~al\mbox{.}(2021)]%
        {abbas2021application}
\bibfield{author}{\bibinfo{person}{Khushnood Abbas}, \bibinfo{person}{Alireza
  Abbasi}, \bibinfo{person}{Shi Dong}, \bibinfo{person}{Ling Niu},
  \bibinfo{person}{Laihang Yu}, \bibinfo{person}{Bolun Chen},
  \bibinfo{person}{Shi-Min Cai}, {and} \bibinfo{person}{Qambar Hasan}.}
  \bibinfo{year}{2021}\natexlab{}.
\newblock \showarticletitle{Application of network link prediction in drug
  discovery}.
\newblock \bibinfo{journal}{\emph{BMC bioinformatics}} \bibinfo{volume}{22},
  \bibinfo{number}{1} (\bibinfo{year}{2021}), \bibinfo{pages}{1--21}.
\newblock


\bibitem[Alemi et~al\mbox{.}(2018)]%
        {alemi2018fixing}
\bibfield{author}{\bibinfo{person}{Alexander Alemi}, \bibinfo{person}{Ben
  Poole}, \bibinfo{person}{Ian Fischer}, \bibinfo{person}{Joshua Dillon},
  \bibinfo{person}{Rif~A Saurous}, {and} \bibinfo{person}{Kevin Murphy}.}
  \bibinfo{year}{2018}\natexlab{}.
\newblock \showarticletitle{Fixing a broken ELBO}. In
  \bibinfo{booktitle}{\emph{International Conference on Machine Learning}}.
  PMLR, \bibinfo{pages}{159--168}.
\newblock


\bibitem[Calvo et~al\mbox{.}(1990)]%
        {calvo1990fifth}
\bibfield{author}{\bibinfo{person}{M Calvo}, \bibinfo{person}{JI Montijano},
  {and} \bibinfo{person}{L Randez}.} \bibinfo{year}{1990}\natexlab{}.
\newblock \showarticletitle{A fifth-order interpolant for the Dormand and
  Prince Runge-Kutta method}.
\newblock \bibinfo{journal}{\emph{Journal of computational and applied
  mathematics}} \bibinfo{volume}{29}, \bibinfo{number}{1}
  (\bibinfo{year}{1990}), \bibinfo{pages}{91--100}.
\newblock


\bibitem[Cangea et~al\mbox{.}(2022)]%
        {cangea2022message}
\bibfield{author}{\bibinfo{person}{C{\u{a}}t{\u{a}}lina Cangea},
  \bibinfo{person}{Ben Day}, \bibinfo{person}{Arian~Rokkum Jamasb}, {and}
  \bibinfo{person}{Pietro Lio}.} \bibinfo{year}{2022}\natexlab{}.
\newblock \showarticletitle{Message Passing Neural Processes}. In
  \bibinfo{booktitle}{\emph{ICLR 2022 Workshop on Geometrical and Topological
  Representation Learning}}.
\newblock


\bibitem[Cao et~al\mbox{.}(2021)]%
        {cao2021inductive}
\bibfield{author}{\bibinfo{person}{Huafeng Cao}, \bibinfo{person}{Zhongbao
  Zhang}, \bibinfo{person}{Li Sun}, {and} \bibinfo{person}{Zhi Wang}.}
  \bibinfo{year}{2021}\natexlab{}.
\newblock \showarticletitle{Inductive and irregular dynamic network
  representation based on ordinary differential equations}.
\newblock \bibinfo{journal}{\emph{Knowledge-Based Systems}}
  \bibinfo{volume}{227} (\bibinfo{year}{2021}), \bibinfo{pages}{107271}.
\newblock


\bibitem[Chen et~al\mbox{.}(2020)]%
        {chen2020neural}
\bibfield{author}{\bibinfo{person}{Ricky~TQ Chen}, \bibinfo{person}{Brandon
  Amos}, {and} \bibinfo{person}{Maximilian Nickel}.}
  \bibinfo{year}{2020}\natexlab{}.
\newblock \showarticletitle{Neural Spatio-Temporal Point Processes}. In
  \bibinfo{booktitle}{\emph{International Conference on Learning
  Representations}}.
\newblock


\bibitem[Chen et~al\mbox{.}(2018)]%
        {chen2018neural}
\bibfield{author}{\bibinfo{person}{Ricky~TQ Chen}, \bibinfo{person}{Yulia
  Rubanova}, \bibinfo{person}{Jesse Bettencourt}, {and}
  \bibinfo{person}{David~K Duvenaud}.} \bibinfo{year}{2018}\natexlab{}.
\newblock \showarticletitle{Neural ordinary differential equations}.
\newblock \bibinfo{journal}{\emph{Advances in neural information processing
  systems}}  \bibinfo{volume}{31} (\bibinfo{year}{2018}).
\newblock


\bibitem[Chung et~al\mbox{.}(2014)]%
        {chung2014empirical}
\bibfield{author}{\bibinfo{person}{Junyoung Chung}, \bibinfo{person}{Caglar
  Gulcehre}, \bibinfo{person}{Kyunghyun Cho}, {and} \bibinfo{person}{Yoshua
  Bengio}.} \bibinfo{year}{2014}\natexlab{}.
\newblock \showarticletitle{Empirical evaluation of gated recurrent neural
  networks on sequence modeling}. In \bibinfo{booktitle}{\emph{NIPS 2014
  Workshop on Deep Learning, December 2014}}.
\newblock


\bibitem[Dong et~al\mbox{.}(2012)]%
        {6413904}
\bibfield{author}{\bibinfo{person}{Yuxiao Dong}, \bibinfo{person}{Jie Tang},
  \bibinfo{person}{Sen Wu}, \bibinfo{person}{Jilei Tian},
  \bibinfo{person}{Nitesh~V. Chawla}, \bibinfo{person}{Jinghai Rao}, {and}
  \bibinfo{person}{Huanhuan Cao}.} \bibinfo{year}{2012}\natexlab{}.
\newblock \showarticletitle{Link Prediction and Recommendation across
  Heterogeneous Social Networks}. In \bibinfo{booktitle}{\emph{2012 IEEE 12th
  International Conference on Data Mining}}. \bibinfo{pages}{181--190}.
\newblock
\urldef\tempurl%
\url{https://doi.org/10.1109/ICDM.2012.140}
\showDOI{\tempurl}


\bibitem[Eslami et~al\mbox{.}(2018)]%
        {eslami2018neural}
\bibfield{author}{\bibinfo{person}{SM~Ali Eslami}, \bibinfo{person}{Danilo
  Jimenez~Rezende}, \bibinfo{person}{Frederic Besse}, \bibinfo{person}{Fabio
  Viola}, \bibinfo{person}{Ari~S Morcos}, \bibinfo{person}{Marta Garnelo},
  \bibinfo{person}{Avraham Ruderman}, \bibinfo{person}{Andrei~A Rusu},
  \bibinfo{person}{Ivo Danihelka}, \bibinfo{person}{Karol Gregor},
  {et~al\mbox{.}}} \bibinfo{year}{2018}\natexlab{}.
\newblock \showarticletitle{Neural scene representation and rendering}.
\newblock \bibinfo{journal}{\emph{Science}} \bibinfo{volume}{360},
  \bibinfo{number}{6394} (\bibinfo{year}{2018}), \bibinfo{pages}{1204--1210}.
\newblock


\bibitem[Garnelo et~al\mbox{.}(2018a)]%
        {garnelo2018conditional}
\bibfield{author}{\bibinfo{person}{Marta Garnelo}, \bibinfo{person}{Dan
  Rosenbaum}, \bibinfo{person}{Christopher Maddison}, \bibinfo{person}{Tiago
  Ramalho}, \bibinfo{person}{David Saxton}, \bibinfo{person}{Murray Shanahan},
  \bibinfo{person}{Yee~Whye Teh}, \bibinfo{person}{Danilo Rezende}, {and}
  \bibinfo{person}{SM~Ali Eslami}.} \bibinfo{year}{2018}\natexlab{a}.
\newblock \showarticletitle{Conditional neural processes}. In
  \bibinfo{booktitle}{\emph{International Conference on Machine Learning}}.
  PMLR, \bibinfo{pages}{1704--1713}.
\newblock


\bibitem[Garnelo et~al\mbox{.}(2018b)]%
        {garnelo2018neural}
\bibfield{author}{\bibinfo{person}{Marta Garnelo}, \bibinfo{person}{Jonathan
  Schwarz}, \bibinfo{person}{Dan Rosenbaum}, \bibinfo{person}{Fabio Viola},
  \bibinfo{person}{Danilo~J Rezende}, \bibinfo{person}{SM Eslami}, {and}
  \bibinfo{person}{Yee~Whye Teh}.} \bibinfo{year}{2018}\natexlab{b}.
\newblock \showarticletitle{Neural processes}.
\newblock \bibinfo{journal}{\emph{arXiv preprint arXiv:1807.01622}}
  (\bibinfo{year}{2018}).
\newblock


\bibitem[Han et~al\mbox{.}(2021)]%
        {han2021learning}
\bibfield{author}{\bibinfo{person}{Zhen Han}, \bibinfo{person}{Zifeng Ding},
  \bibinfo{person}{Yunpu Ma}, \bibinfo{person}{Yujia Gu}, {and}
  \bibinfo{person}{Volker Tresp}.} \bibinfo{year}{2021}\natexlab{}.
\newblock \showarticletitle{Learning neural ordinary equations for forecasting
  future links on temporal knowledge graphs}. In
  \bibinfo{booktitle}{\emph{Proceedings of the 2021 Conference on Empirical
  Methods in Natural Language Processing}}. \bibinfo{pages}{8352--8364}.
\newblock


\bibitem[Holderrieth et~al\mbox{.}(2021)]%
        {holderrieth2021equivariant}
\bibfield{author}{\bibinfo{person}{Peter Holderrieth},
  \bibinfo{person}{Michael~J Hutchinson}, {and} \bibinfo{person}{Yee~Whye
  Teh}.} \bibinfo{year}{2021}\natexlab{}.
\newblock \showarticletitle{Equivariant learning of stochastic fields: Gaussian
  processes and steerable conditional neural processes}. In
  \bibinfo{booktitle}{\emph{International Conference on Machine Learning}}.
  PMLR, \bibinfo{pages}{4297--4307}.
\newblock


\bibitem[Huo et~al\mbox{.}(2018)]%
        {huo2018link}
\bibfield{author}{\bibinfo{person}{Zepeng Huo}, \bibinfo{person}{Xiao Huang},
  {and} \bibinfo{person}{Xia Hu}.} \bibinfo{year}{2018}\natexlab{}.
\newblock \showarticletitle{Link prediction with personalized social
  influence}. In \bibinfo{booktitle}{\emph{Proceedings of the AAAI Conference
  on Artificial Intelligence}}, Vol.~\bibinfo{volume}{32}.
\newblock


\bibitem[Jin et~al\mbox{.}(2022)]%
        {jin2022neural}
\bibfield{author}{\bibinfo{person}{Ming Jin}, \bibinfo{person}{Yuan-Fang Li},
  {and} \bibinfo{person}{Shirui Pan}.} \bibinfo{year}{2022}\natexlab{}.
\newblock \showarticletitle{Neural Temporal Walks: Motif-Aware Representation
  Learning on Continuous-Time Dynamic Graphs}. In
  \bibinfo{booktitle}{\emph{36th Conference on Neural Information Processing
  Systems}}.
\newblock


\bibitem[Kim et~al\mbox{.}(2018)]%
        {kim2018attentive}
\bibfield{author}{\bibinfo{person}{Hyunjik Kim}, \bibinfo{person}{Andriy Mnih},
  \bibinfo{person}{Jonathan Schwarz}, \bibinfo{person}{Marta Garnelo},
  \bibinfo{person}{Ali Eslami}, \bibinfo{person}{Dan Rosenbaum},
  \bibinfo{person}{Oriol Vinyals}, {and} \bibinfo{person}{Yee~Whye Teh}.}
  \bibinfo{year}{2018}\natexlab{}.
\newblock \showarticletitle{Attentive Neural Processes}. In
  \bibinfo{booktitle}{\emph{International Conference on Learning
  Representations}}.
\newblock


\bibitem[Kovanen et~al\mbox{.}(2011)]%
        {kovanen2011temporal}
\bibfield{author}{\bibinfo{person}{Lauri Kovanen}, \bibinfo{person}{M{\'a}rton
  Karsai}, \bibinfo{person}{Kimmo Kaski}, \bibinfo{person}{J{\'a}nos
  Kert{\'e}sz}, {and} \bibinfo{person}{Jari Saram{\"a}ki}.}
  \bibinfo{year}{2011}\natexlab{}.
\newblock \showarticletitle{Temporal motifs in time-dependent networks}.
\newblock \bibinfo{journal}{\emph{Journal of Statistical Mechanics: Theory and
  Experiment}} \bibinfo{volume}{2011}, \bibinfo{number}{11}
  (\bibinfo{year}{2011}), \bibinfo{pages}{P11005}.
\newblock


\bibitem[Kumar et~al\mbox{.}(2019)]%
        {kumar2019predicting}
\bibfield{author}{\bibinfo{person}{Srijan Kumar}, \bibinfo{person}{Xikun
  Zhang}, {and} \bibinfo{person}{Jure Leskovec}.}
  \bibinfo{year}{2019}\natexlab{}.
\newblock \showarticletitle{Predicting dynamic embedding trajectory in temporal
  interaction networks}. In \bibinfo{booktitle}{\emph{Proceedings of the 25th
  ACM SIGKDD international conference on knowledge discovery \& data mining}}.
  \bibinfo{pages}{1269--1278}.
\newblock


\bibitem[Liang and Gao(2022)]%
        {liang2022neural}
\bibfield{author}{\bibinfo{person}{Huidong Liang} {and} \bibinfo{person}{Junbin
  Gao}.} \bibinfo{year}{2022}\natexlab{}.
\newblock \showarticletitle{How Neural Processes Improve Graph Link
  Prediction}. In \bibinfo{booktitle}{\emph{ICASSP 2022-2022 IEEE International
  Conference on Acoustics, Speech and Signal Processing (ICASSP)}}. IEEE,
  \bibinfo{pages}{3543--3547}.
\newblock


\bibitem[Lin et~al\mbox{.}(2022)]%
        {lin2022distributed}
\bibfield{author}{\bibinfo{person}{Haiyang Lin}, \bibinfo{person}{Mingyu Yan},
  \bibinfo{person}{Xiaochun Ye}, \bibinfo{person}{Dongrui Fan},
  \bibinfo{person}{Shirui Pan}, \bibinfo{person}{Wenguang Chen}, {and}
  \bibinfo{person}{Yuan Xie}.} \bibinfo{year}{2022}\natexlab{}.
\newblock \bibinfo{title}{A Comprehensive Survey on Distributed Training of
  Graph Neural Networks}.
\newblock
\newblock
\urldef\tempurl%
\url{https://doi.org/10.48550/ARXIV.2211.05368}
\showDOI{\tempurl}


\bibitem[Lin et~al\mbox{.}(2021)]%
        {lin2021task}
\bibfield{author}{\bibinfo{person}{Xixun Lin}, \bibinfo{person}{Jia Wu},
  \bibinfo{person}{Chuan Zhou}, \bibinfo{person}{Shirui Pan},
  \bibinfo{person}{Yanan Cao}, {and} \bibinfo{person}{Bin Wang}.}
  \bibinfo{year}{2021}\natexlab{}.
\newblock \showarticletitle{Task-adaptive neural process for user cold-start
  recommendation}. In \bibinfo{booktitle}{\emph{Proceedings of the Web
  Conference 2021}}. \bibinfo{pages}{1306--1316}.
\newblock


\bibitem[Liu et~al\mbox{.}(2021)]%
        {liu2021anomaly}
\bibfield{author}{\bibinfo{person}{Yixin Liu}, \bibinfo{person}{Zhao Li},
  \bibinfo{person}{Shirui Pan}, \bibinfo{person}{Chen Gong},
  \bibinfo{person}{Chuan Zhou}, {and} \bibinfo{person}{George Karypis}.}
  \bibinfo{year}{2021}\natexlab{}.
\newblock \showarticletitle{Anomaly detection on attributed networks via
  contrastive self-supervised learning}.
\newblock \bibinfo{journal}{\emph{IEEE transactions on neural networks and
  learning systems}} \bibinfo{volume}{33}, \bibinfo{number}{6}
  (\bibinfo{year}{2021}), \bibinfo{pages}{2378--2392}.
\newblock


\bibitem[Luo et~al\mbox{.}(2021)]%
        {luo2021detecting}
\bibfield{author}{\bibinfo{person}{Linhao Luo}, \bibinfo{person}{Yixiang Fang},
  \bibinfo{person}{Xin Cao}, \bibinfo{person}{Xiaofeng Zhang}, {and}
  \bibinfo{person}{Wenjie Zhang}.} \bibinfo{year}{2021}\natexlab{}.
\newblock \showarticletitle{Detecting communities from heterogeneous graphs: A
  context path-based graph neural network model}. In
  \bibinfo{booktitle}{\emph{Proceedings of the 30th ACM International
  Conference on Information \& Knowledge Management}}.
  \bibinfo{pages}{1170--1180}.
\newblock


\bibitem[Nguyen and Grover(2022)]%
        {nguyen2022transformer}
\bibfield{author}{\bibinfo{person}{Tung Nguyen} {and} \bibinfo{person}{Aditya
  Grover}.} \bibinfo{year}{2022}\natexlab{}.
\newblock \showarticletitle{Transformer Neural Processes: Uncertainty-Aware
  Meta Learning Via Sequence Modeling}. In
  \bibinfo{booktitle}{\emph{International Conference on Machine Learning}}.
  PMLR, \bibinfo{pages}{16569--16594}.
\newblock


\bibitem[Norcliffe et~al\mbox{.}(2020)]%
        {norcliffe2020neural}
\bibfield{author}{\bibinfo{person}{Alexander Norcliffe},
  \bibinfo{person}{Cristian Bodnar}, \bibinfo{person}{Ben Day},
  \bibinfo{person}{Jacob Moss}, {and} \bibinfo{person}{Pietro Li{\`o}}.}
  \bibinfo{year}{2020}\natexlab{}.
\newblock \showarticletitle{Neural ODE Processes}. In
  \bibinfo{booktitle}{\emph{International Conference on Learning
  Representations}}.
\newblock


\bibitem[Paranjape et~al\mbox{.}(2017)]%
        {paranjape2017motifs}
\bibfield{author}{\bibinfo{person}{Ashwin Paranjape}, \bibinfo{person}{Austin~R
  Benson}, {and} \bibinfo{person}{Jure Leskovec}.}
  \bibinfo{year}{2017}\natexlab{}.
\newblock \showarticletitle{Motifs in temporal networks}. In
  \bibinfo{booktitle}{\emph{Proceedings of the tenth ACM international
  conference on web search and data mining}}. \bibinfo{pages}{601--610}.
\newblock


\bibitem[Pareja et~al\mbox{.}(2020)]%
        {pareja2020evolvegcn}
\bibfield{author}{\bibinfo{person}{Aldo Pareja}, \bibinfo{person}{Giacomo
  Domeniconi}, \bibinfo{person}{Jie Chen}, \bibinfo{person}{Tengfei Ma},
  \bibinfo{person}{Toyotaro Suzumura}, \bibinfo{person}{Hiroki Kanezashi},
  \bibinfo{person}{Tim Kaler}, \bibinfo{person}{Tao Schardl}, {and}
  \bibinfo{person}{Charles Leiserson}.} \bibinfo{year}{2020}\natexlab{}.
\newblock \showarticletitle{Evolvegcn: Evolving graph convolutional networks
  for dynamic graphs}. In \bibinfo{booktitle}{\emph{Proceedings of the AAAI
  Conference on Artificial Intelligence}}, Vol.~\bibinfo{volume}{34}.
  \bibinfo{pages}{5363--5370}.
\newblock


\bibitem[Rossi et~al\mbox{.}(2020)]%
        {tgn_icml_grl2020}
\bibfield{author}{\bibinfo{person}{Emanuele Rossi}, \bibinfo{person}{Ben
  Chamberlain}, \bibinfo{person}{Fabrizio Frasca}, \bibinfo{person}{Davide
  Eynard}, \bibinfo{person}{Federico Monti}, {and} \bibinfo{person}{Michael
  Bronstein}.} \bibinfo{year}{2020}\natexlab{}.
\newblock \showarticletitle{Temporal Graph Networks for Deep Learning on
  Dynamic Graphs}. In \bibinfo{booktitle}{\emph{ICML 2020 Workshop on Graph
  Representation Learning}}.
\newblock


\bibitem[Rubanova et~al\mbox{.}(2019)]%
        {rubanova2019latent}
\bibfield{author}{\bibinfo{person}{Yulia Rubanova}, \bibinfo{person}{Ricky~TQ
  Chen}, {and} \bibinfo{person}{David~K Duvenaud}.}
  \bibinfo{year}{2019}\natexlab{}.
\newblock \showarticletitle{Latent ordinary differential equations for
  irregularly-sampled time series}.
\newblock \bibinfo{journal}{\emph{Advances in neural information processing
  systems}}  \bibinfo{volume}{32} (\bibinfo{year}{2019}).
\newblock


\bibitem[Sankar et~al\mbox{.}(2020)]%
        {sankar2020dysat}
\bibfield{author}{\bibinfo{person}{Aravind Sankar}, \bibinfo{person}{Yanhong
  Wu}, \bibinfo{person}{Liang Gou}, \bibinfo{person}{Wei Zhang}, {and}
  \bibinfo{person}{Hao Yang}.} \bibinfo{year}{2020}\natexlab{}.
\newblock \showarticletitle{Dysat: Deep neural representation learning on
  dynamic graphs via self-attention networks}. In
  \bibinfo{booktitle}{\emph{Proceedings of the 13th international conference on
  web search and data mining}}. \bibinfo{pages}{519--527}.
\newblock


\bibitem[Singh et~al\mbox{.}(2019)]%
        {singh2019sequential}
\bibfield{author}{\bibinfo{person}{Gautam Singh}, \bibinfo{person}{Jaesik
  Yoon}, \bibinfo{person}{Youngsung Son}, {and} \bibinfo{person}{Sungjin Ahn}.}
  \bibinfo{year}{2019}\natexlab{}.
\newblock \showarticletitle{Sequential neural processes}.
\newblock \bibinfo{journal}{\emph{Advances in Neural Information Processing
  Systems}}  \bibinfo{volume}{32} (\bibinfo{year}{2019}).
\newblock


\bibitem[Wang et~al\mbox{.}(2021b)]%
        {wang2021apan}
\bibfield{author}{\bibinfo{person}{Xuhong Wang}, \bibinfo{person}{Ding Lyu},
  \bibinfo{person}{Mengjian Li}, \bibinfo{person}{Yang Xia},
  \bibinfo{person}{Qi Yang}, \bibinfo{person}{Xinwen Wang},
  \bibinfo{person}{Xinguang Wang}, \bibinfo{person}{Ping Cui},
  \bibinfo{person}{Yupu Yang}, \bibinfo{person}{Bowen Sun}, {et~al\mbox{.}}}
  \bibinfo{year}{2021}\natexlab{b}.
\newblock \showarticletitle{APAN: Asynchronous propagation attention network
  for real-time temporal graph embedding}. In
  \bibinfo{booktitle}{\emph{Proceedings of the 2021 International Conference on
  Management of Data}}. \bibinfo{pages}{2628--2638}.
\newblock


\bibitem[Wang et~al\mbox{.}(2021a)]%
        {wang2021inductive}
\bibfield{author}{\bibinfo{person}{Yanbang Wang}, \bibinfo{person}{Yen-Yu
  Chang}, \bibinfo{person}{Yunyu Liu}, \bibinfo{person}{Jure Leskovec}, {and}
  \bibinfo{person}{Pan Li}.} \bibinfo{year}{2021}\natexlab{a}.
\newblock \showarticletitle{Inductive Representation Learning in Temporal
  Networks via Causal Anonymous Walks}. In
  \bibinfo{booktitle}{\emph{International Conference on Learning
  Representations (ICLR)}}.
\newblock


\bibitem[Welling and Kipf(2016)]%
        {welling2016semi}
\bibfield{author}{\bibinfo{person}{Max Welling} {and} \bibinfo{person}{Thomas~N
  Kipf}.} \bibinfo{year}{2016}\natexlab{}.
\newblock \showarticletitle{Semi-supervised classification with graph
  convolutional networks}. In \bibinfo{booktitle}{\emph{J. International
  Conference on Learning Representations (ICLR 2017)}}.
\newblock


\bibitem[Wen and Fang(2022)]%
        {wen2022trend}
\bibfield{author}{\bibinfo{person}{Zhihao Wen} {and} \bibinfo{person}{Yuan
  Fang}.} \bibinfo{year}{2022}\natexlab{}.
\newblock \showarticletitle{TREND: TempoRal Event and Node Dynamics for Graph
  Representation Learning}. In \bibinfo{booktitle}{\emph{Proceedings of the ACM
  Web Conference 2022}}. \bibinfo{pages}{1159--1169}.
\newblock


\bibitem[Xiong et~al\mbox{.}(2022)]%
        {xiong2022pseudo}
\bibfield{author}{\bibinfo{person}{Bo Xiong}, \bibinfo{person}{Shichao Zhu},
  \bibinfo{person}{Nico Potyka}, \bibinfo{person}{Shirui Pan},
  \bibinfo{person}{Chuan Zhou}, {and} \bibinfo{person}{Steffen Staab}.}
  \bibinfo{year}{2022}\natexlab{}.
\newblock \showarticletitle{Pseudo-Riemannian Graph Convolutional Networks}. In
  \bibinfo{booktitle}{\emph{36th Conference on Neural Information Processing
  Systems}}.
\newblock


\bibitem[Xu et~al\mbox{.}(2020)]%
        {xu2020inductive}
\bibfield{author}{\bibinfo{person}{Da Xu}, \bibinfo{person}{Chuanwei Ruan},
  \bibinfo{person}{Evren Korpeoglu}, \bibinfo{person}{Sushant Kumar}, {and}
  \bibinfo{person}{Kannan Achan}.} \bibinfo{year}{2020}\natexlab{}.
\newblock \showarticletitle{Inductive representation learning on temporal
  graphs}. In \bibinfo{booktitle}{\emph{ICLR 2020}}.
\newblock


\bibitem[Yang et~al\mbox{.}(2022)]%
        {yang2022few}
\bibfield{author}{\bibinfo{person}{Cheng Yang}, \bibinfo{person}{Chunchen
  Wang}, \bibinfo{person}{Yuanfu Lu}, \bibinfo{person}{Xumeng Gong},
  \bibinfo{person}{Chuan Shi}, \bibinfo{person}{Wei Wang}, {and}
  \bibinfo{person}{Xu Zhang}.} \bibinfo{year}{2022}\natexlab{}.
\newblock \showarticletitle{Few-shot Link Prediction in Dynamic Networks}. In
  \bibinfo{booktitle}{\emph{Proceedings of the Fifteenth ACM International
  Conference on Web Search and Data Mining}}. \bibinfo{pages}{1245--1255}.
\newblock


\bibitem[Yokoi et~al\mbox{.}(2017)]%
        {yokoi2017link}
\bibfield{author}{\bibinfo{person}{Sho Yokoi}, \bibinfo{person}{Hiroshi
  Kajino}, {and} \bibinfo{person}{Hisashi Kashima}.}
  \bibinfo{year}{2017}\natexlab{}.
\newblock \showarticletitle{Link prediction in sparse networks by incidence
  matrix factorization}.
\newblock \bibinfo{journal}{\emph{Journal of Information Processing}}
  \bibinfo{volume}{25} (\bibinfo{year}{2017}), \bibinfo{pages}{477--485}.
\newblock


\bibitem[Zheng et~al\mbox{.}(2022b)]%
        {zheng2022multi}
\bibfield{author}{\bibinfo{person}{Xin Zheng}, \bibinfo{person}{Miao Zhang},
  \bibinfo{person}{Chunyang Chen}, \bibinfo{person}{Chaojie Li},
  \bibinfo{person}{Chuan Zhou}, {and} \bibinfo{person}{Shirui Pan}.}
  \bibinfo{year}{2022}\natexlab{b}.
\newblock \showarticletitle{Multi-Relational Graph Neural Architecture Search
  with Fine-grained Message Passing}. In \bibinfo{booktitle}{\emph{22nd IEEE
  International Conference on Data Mining}}.
\newblock


\bibitem[Zheng et~al\mbox{.}(2022a)]%
        {zheng2022rethink}
\bibfield{author}{\bibinfo{person}{Yizhen Zheng}, \bibinfo{person}{Shirui Pan},
  \bibinfo{person}{Vincent Lee}, \bibinfo{person}{Yu Zheng}, {and}
  \bibinfo{person}{Philip~S. Yu}.} \bibinfo{year}{2022}\natexlab{a}.
\newblock \showarticletitle{Rethinking and Scaling Up Graph Contrastive
  Learning: An Extremely Efficient Approach with Group Discrimination}. In
  \bibinfo{booktitle}{\emph{36th Conference on Neural Information Processing
  Systems}}.
\newblock


\end{thebibliography}


\end{document}


\maketitle
\section{Derivation of ELBO Loss}
Given the context data $G_{\leq T}$ and target data $G_{> T}$, the object of \ourmethod is to infer the distribution $P(z_{>T}|G_{\leq T})$ from the context data that minimize the prediction loss on the target data. Therefore, the generation process of \ourmethod can be written as 
\begin{equation}
    \label{eq:gen}
    P(z_{>T},Y_D|\mathcal{E}_D,\mathcal{G}_{\leq T}) = P(z_{>T}|G_{\leq T}) \prod_{e_D(t)\in \mathcal{E}_D} P\big(y_D|f(e_D(t),z_{>T})\big),
\end{equation}
where $f$ is the link prediction decoder, and $\mathcal{E}_D=\{e_D(t)\in G_{>T}\}$ denotes the links to be predicted in the target data.

Following the Equation \ref{eq:gen}, we can rewrite the $logP(Y_D|\mathcal{E}_D,G_{\leq T})$ as 
\begin{align}
    logP(Y_D|\mathcal{E}_D,G_{\leq T}) &= log\frac{P(Y_D, z_{>T}|\mathcal{E}_D,G_{\leq T})}{P(z_{>T}|G_{\leq T})},\\
    \label{eq:log}
    &=logP(Y_D, z_{>T}|\mathcal{E}_D,G_{\leq T}) - logP(z_{>T}|G_{\leq T}).
\end{align}

Assuming the true distribution of $z_{>T}$ is $Q(z_{>T})$, we can rewrite the Equation \ref{eq:log} as
\begin{equation}
    logP(Y_D|\mathcal{E}_D,G_{\leq T}) = log\frac{P(Y_D, z_{>T}|\mathcal{E}_D,G_{\leq T})}{Q(z_{>T})} - log\frac{P(z_{>T}|G_{\leq T})}{Q(z_{>T})}.
\end{equation}

We integrate both side with $Q(z_{>T})$.
\begin{align}
    \int_{z}Q(z_{>T}) logP(Y_D|\mathcal{E}_D,G_{\leq T}) &= \int_{z}Q(z_{>T})log\frac{P(Y_D, z_{>T}|\mathcal{E}_D,G_{\leq T})}{Q(z_{>T})} - \int_{z}Q(z_{>T})log\frac{P(z_{>T}|G_{\leq T})}{Q(z_{>T})},\\
    \label{eq:int}
    logP(Y_D|\mathcal{E}_D,G_{\leq T}) &=
        \int_{z}Q(z_{>T})log\frac{P(Y_D, z_{>T}|\mathcal{E}_D,G_{\leq T})}{Q(z_{>T})} + KL\big(Q(z_{>T})||P(z_{>T}|G_{\leq T})\big).
\end{align}

Because $KL\big(Q(z_{>T})||P(z_{>T}|G_{\leq T})\big) \geq 0$, we can write Equation \ref{eq:int} as
\begin{align}
    logP(Y_D|\mathcal{E}_D,G_{\leq T}) &\geq \int_{z}Q(z_{>T})log\frac{P(Y_D, z_{>T}|\mathcal{E}_D,G_{\leq T})}{Q(z_{>T})}, \\
    &= \mathbb{E}_{Q(z_{>T})}log\frac{P(Y_D, z_{>T}|\mathcal{E}_D,G_{\leq T})}{Q(z_{>T})},\\
    &= \mathbb{E}_{Q(z_{>T})}[logP(Y_D|\mathcal{E}_D,z_{>T}) + log\frac{P(z_{>T}|G_{\leq T})}{Q(z_{>T})}],\\
    &= \mathbb{E}_{Q(z_{>T})}[logP(Y_D|\mathcal{E}_D,z_{>T})] - KL\big(Q(z_{>T})||P(z_{>T}|G_{\leq T})\big)\label{eq:last},
\end{align}
where $P(Y_D|\mathcal{E}_D,z_{>T})$ is calculated by the link prediction decoder $P_\phi(Y_D|\mathcal{E}_D,z_{>T})$, $Q(z_{>T})$ is approximated by the encoder and RNN aggregator, written as $Q_\psi(z|G_{\leq T},G_{> T})$, and $P(z_{>T}|G_{\leq T})$ is approximated by the NODE, written as $P_{ode}(z_{>T}|G_{\leq T})$.

Thus, we can obtain the final ELBO loss as
\begin{equation}
    \mathcal{L} = \mathbb{E}_{Q_\psi(z_{>T}|G_{\leq T},G_{> T})}[log P_\phi(Y_D|\mathcal{E}_D,z_{>T})]
    - KL(Q_\psi(z_{>T}|G_{\leq T},G_{> T})||P_{ode}(z_{>T}|G_{\leq T})).
\end{equation}

\section{Optimization of ELBO Loss}

To optimize ELBO loss, we need to compute the gradients with respect to the parameters of the model.

In ELBO loss, we encourage the NODE to learn the derivative of distribution and infer the distribution in the future by minimizing the KL divergence between variational posterior $Q_\psi(z_T|G_{\leq T},G_{> T})$ and prior $P_{ode}(z_{>T}|G_{\leq T})$.
Thus, the gradients for $\psi$ and NODE can be directly computed.

However, for the first expectation term, since $z_{>T}$ is sampled from $\mathcal{N}\big(\mu(\mathbf{r'}_{>T}),\sigma(\mathbf{r'}_{>T})\big)$, we adopt the reparameterization trick \cite{kingma2013auto} to compute the gradients and compute the expectation using Monte Carlo methods, written as
\begin{gather}
    z_{>T}^l = \mu(\mathbf{r'}_{>T}) + \sigma(\mathbf{r'}_{>T})\epsilon^l, \epsilon^l\sim \mathcal{N}(0,1),\\ 
    \mathbb{E}_{Q(z_{>T})}[logP_\phi(Y_D|\mathcal{E}_D,z_{>T})] \simeq \frac{1}{L} \sum_{l=1}^L logP_\phi(Y_D|\mathcal{E}_D,z_{>T}^l).
\end{gather}
We set $L=10$ in experiments.

For parameters $\theta$ in NODE, we adopt the adjoint sensitivity method \cite{pontryagin1987mathematical,chen2018neural} to compute the gradients $\frac{d\mathcal{L}}{d\theta}$, written as
\begin{gather}
    \frac{d\mathcal{L}}{d\theta} = -\int_{T+\triangle T}^{T} \mathbf{a}(t)^{\top}\frac{\partial f_{ode}(\mathbf{r}_t,t,\theta)}{\partial \theta},
\end{gather}
where $\mathbf{a}(t)$ is denoted as the gradient of the loss depends on the hidden state. Its dynamics are given by another ODE, written as
\begin{equation}
    \frac{d\mathbf{a}(t)}{dt} = - \mathbf{a}(t)^{\top}\frac{\partial f_{ode}(\mathbf{r}_t,t,\theta)}{\partial \mathbf{r}_t}.
\end{equation}


\bibliographystyle{ACM-Reference-Format}
\bibliography{sections/ref.bib}